\PassOptionsToPackage{table}{xcolor}
\documentclass{article} 
\usepackage{iclr2025_conference,times}

\usepackage{amsmath,amsfonts,bm}

\def\eqref#1{equation~\ref{#1}}

\def\1{\bm{1}}

\DeclareMathAlphabet{\mathsfit}{\encodingdefault}{\sfdefault}{m}{sl}
\SetMathAlphabet{\mathsfit}{bold}{\encodingdefault}{\sfdefault}{bx}{n}

\usepackage{url}
\usepackage{booktabs}
\usepackage{multirow}
\usepackage[table]{xcolor}
\usepackage{graphicx}
\usepackage{amsmath}
\usepackage {amssymb}
\usepackage{graphicx}
\usepackage{wrapfig}
\usepackage{needspace}
\usepackage{algorithm}
\usepackage{algpseudocode}
\usepackage{float}
\usepackage[most]{tcolorbox}
\tcbuselibrary{breakable,skins}
\usepackage{caption}

\usepackage{tabularx}
\usepackage{array}

\usepackage{hyperref}

\definecolor{linkcol}{RGB}{31,102,168}  
\definecolor{citecol}{RGB}{30,130,95}   
\definecolor{urlcol}{RGB}{140,50,120}   

\hypersetup{
    colorlinks=true,
    linkcolor=linkcol,
    citecolor=citecol,
    urlcolor=urlcol,
    filecolor=urlcol,
    linktocpage=true,   
    pdfborder={0 0 0}
}

\usepackage{etoc}

\title{SIVA-RL: Sensitivity–Invariance Visual Alignment for Multimodal Reinforcement Learning}

\author{%
Cheng Tang$^{1}$\quad
Junzhi Ning$^{1}$\quad
Min Cen$^{1}$\quad
Wei Li$^{1,2}$\quad
Xinyi Zeng$^{1,3}$\quad
Pinxian Zeng$^{5}$\quad
Rongbin Li$^{1}$\quad \\
\textbf{Qiming Zhu$^{4}$}\quad
\textbf{Dongzhan Zhou$^{1}$}\quad
\textbf{Yuqiang Li$^{1}$}\quad 
\textbf{Junjun He$^{1}$}\quad 
\textbf{Yirong Chen$^{1}$}\quad 
\textbf{Ming Hu$^{1}$}\quad
\\
$^{1}$Shanghai Artificial Intelligence Laboratory \quad 
$^{2}$Shanghai Jiao Tong University \quad
$^{3}$Sichuan University \quad \\
$^{4}$The Chinese University of Hong Kong, Shenzhen \quad
$^{5}$University of Macau \quad
}

\iclrpreprintcopy 
\begin{document}

\maketitle

\etocdepthtag.toc{mainbody}

\begin{abstract}
Reinforcement learning with verifiable rewards (RLVR) drives multimodal reasoning, but answer-level correctness does not guarantee that a vision-language model grounds its predictions in visual evidence. Existing visual-intervention methods contrast policy behavior on original and modified images, yet assign supervision by the type of intervention rather than its observed effect. This assumption fails: identical operators produce heterogeneous outcomes across samples. We propose \textbf{SIVA-RL}, a \textbf{S}ensitivity-\textbf{I}nvariance \textbf{V}isual \textbf{A}lignment framework that replaces operator-conditioned regularization with sample-wise, outcome-conditioned supervision. SIVA-RL constructs localized interventions through token-aligned, distance-constrained within-image PatchSwap. A frozen audit policy then scores each clean–intervention pair, and the observed reward drop becomes soft routing weights. Large-drop pairs drive sensitivity alignment, low-drop pairs drive clean-anchored invariance alignment, and ambiguous pairs are down-weighted. This design decouples intervention construction from supervision assignment and is compatible with both GRPO and DAPO backbones. Across nine multimodal reasoning benchmarks spanning mathematical, logical, and vision-dependent tasks, SIVA-RL improves 3B and 7B models over matched RL baselines in every setting. It yields an 8.79 percentage-point gain on vision-dependent reasoning and up to 14.9\% relative overall improvement across all four GRPO- and DAPO-based configurations. \href{https://github.com/tchenglv520/SIVA-RL}{code}


\end{abstract}

\section{Introduction}
\label{sec:introduction}

Reinforcement learning with verifiable rewards (RLVR) improves the reasoning
capabilities of large language and vision-language models
\cite{shao2024deepseekmath,guo2025deepseek,huang2025vision,
yao2025r1,meng2025mm}. By optimizing task-level correctness, RLVR elicits
multi-step reasoning without densely annotated reasoning trajectories.
Final-answer rewards indicate whether a prediction is correct, but not how it
was obtained. A vision-language model can therefore receive high reward while
relying on language priors, answer-frequency biases, or memorized reasoning
templates rather than the visual evidence the task requires
\cite{goyal2017making,agrawal2018don,niu2021counterfactual,
xia2025visionary}. This reward ambiguity is acute in vision-dependent tasks
such as counting, geometry, and diagram understanding.

Recent multimodal RL methods strengthen visual dependence through auxiliary
perception signals, visual credit assignment, or comparisons between clean and
visually modified inputs
\cite{xiao2025perception,huang2025spotlight,wang2026visually,
li2026prpo}. Perturbation-driven methods, most prominently PAPO
\cite{wang2025perception}, regularize the policy by contrasting its behavior
under original and corrupted images. A common design ties the intervention
type to a predetermined objective: destructive interventions encourage
separation, whereas mild or semantics-preserving transformations encourage
consistency
\cite{liu2026noisyrollout,li2025vision,zhang2025cf,
gao2026thinking}. This operator-conditioned design assumes that the same
visual operator induces a sufficiently similar effect across samples.

\begin{figure}[t]
\centering
\includegraphics[width=0.99\textwidth]{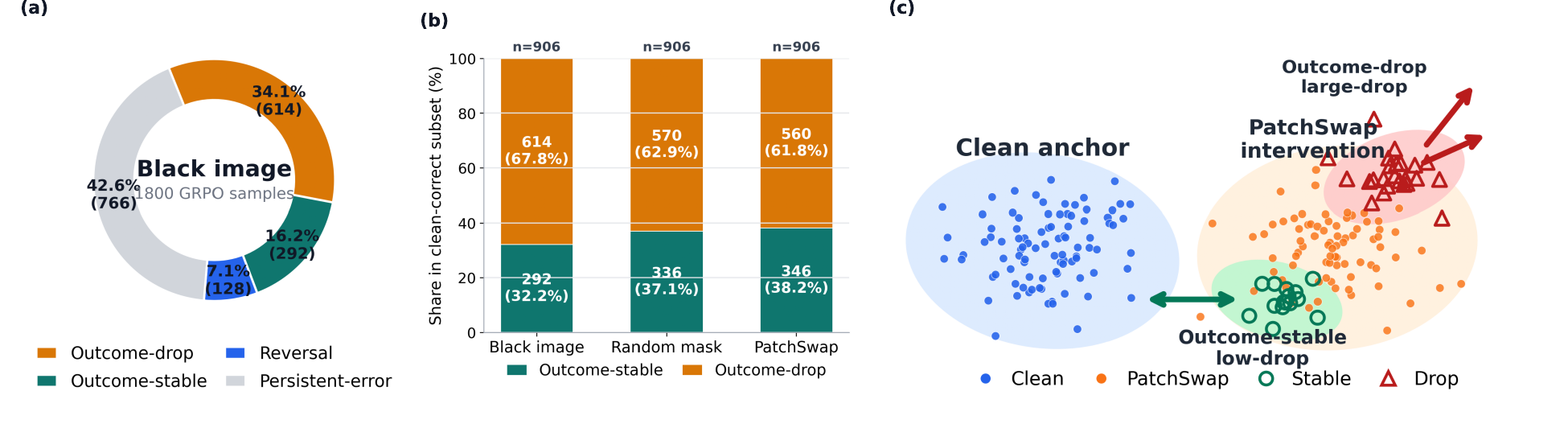}
\caption{
\textbf{Paired intervention audit and SIVA-RL overview.}
\textbf{(a)} Visual interventions induce four paired-outcome categories.
\textbf{(b)} Black-image, random-mask, and PatchSwap interventions all
produce mixtures of outcome-drop and outcome-stable cases among the 906
clean-correct examples.
\textbf{(c)} SIVA-RL routes each valid pair toward sensitivity or
clean-anchored consistency according to its observed outcome change.
}
\label{fig:siva_teaser}
\vspace{-18pt}
\end{figure}

Our paired intervention audit reveals \emph{operator--effect heterogeneity}
at the sample level. We sample 200 examples from each of the nine benchmarks
in our main evaluation, yielding 1,800 examples. For each clean image and its
intervention view, we categorize the answer transition as
\emph{outcome-drop}, \emph{outcome-stable}, \emph{reversal}, or
\emph{persistent-error}. Among the 906 examples answered correctly on their
clean images, black-image, random-mask, and PatchSwap interventions produce
outcome-drop rates of 67.8\%, 62.9\%, and 61.8\%. The corresponding
outcome-stable rates remain high at 32.2\%, 37.1\%, and 38.2\%. The same
operator thus produces both drop and stable outcomes within one clean-correct
subset. Intervention identity alone cannot prescribe a uniform optimization
direction (Figure~\ref{fig:siva_teaser}).

The audit supports one conclusion: paired outcomes are useful for assigning
perturbation-based supervision, but they do not identify the visual evidence
responsible for a prediction. A large reward drop indicates that the policy
behavior is sensitive to the intervention. The change may arise from
task-relevant evidence or from intervention-induced artifacts. A stable
outcome may instead reflect irrelevant modification, redundant visual
evidence, or reliance on a language prior
\cite{goyal2017making,agrawal2018don}. We therefore treat paired outcome
changes as operational routing signals, not certificates of semantic
invariance, causal attribution, or visual grounding.

This heterogeneity raises the central question of this work: \emph{how should
clean--intervention pairs receive different optimization directions and
strengths when the same operator produces heterogeneous sample-level
outcomes?} Our key idea separates intervention construction from the
determination of its training effect. Large-drop pairs provide sensitivity
signals and receive a margin-bounded separation objective. Low-drop pairs
provide outcome-stability signals and receive a conservative, clean-anchored
consistency objective. Ambiguous pairs receive little or no auxiliary
supervision. A clean-correctness gate anchors alignment to a correct clean
response, though the gate does not certify that the response came from visual
evidence.

Building on this principle, we introduce \textbf{SIVA-RL}, a
\textbf{S}ensitivity--\textbf{I}nvariance \textbf{V}isual
\textbf{A}lignment framework for multimodal reinforcement learning.
SIVA-RL decomposes intervention-guided supervision into three stages:
\emph{intervention construction}, \emph{sample-wise outcome valuation}, and
\emph{policy alignment}. SIVA-RL first constructs localized intervention views
through token-aligned, within-image PatchSwap. PatchSwap modifies local visual
content and spatial relationships while preserving most unmodified regions and
the overall image context. An audit policy then evaluates each
clean--intervention pair, and its clean-anchored reward change determines soft
routing weights for a dual-direction alignment objective. Reliable large-drop
pairs are separated, valid low-drop pairs are softly aligned with their clean counterparts, and ambiguous pairs are de-emphasized. Here, ``invariance''
denotes bounded, intervention-conditioned policy consistency, not semantic
invariance or causal irrelevance of the modified content. SIVA-RL thus shifts
visual-intervention-based multimodal RL from operator-conditioned
regularization to sample-wise, outcome-conditioned selective supervision.

We evaluate SIVA-RL on GRPO- and DAPO-based multimodal RL backbones across
model scales and nine benchmarks spanning mathematical, logical, counting,
multi-disciplinary, and vision-dependent reasoning. SIVA-RL improves its
corresponding RL backbones in every configuration. Controlled ablations
confirm the utility of outcome-conditioned routing and the complementary roles
of the sensitivity and invariance branches.

Our main contributions are summarized as follows:
\begin{itemize}
\item \textbf{Diagnosing ambiguity in intervention-defined supervision.}
We show that the same visual intervention can induce mixed outcome-drop and outcome-stable behavior within clean-correct examples. Moreover, low-drop pairs are not necessarily reliable invariance signals, as they may arise from task-irrelevant perturbations, redundant evidence, or language-prior shortcuts. This finding motivates moving beyond operator-defined supervision toward sample-wise effect estimation.

\item \textbf{Outcome-conditioned selective visual alignment.}
We formulate visual-intervention-based multimodal RL as a selective supervision problem. The observed clean-to-intervention reward drop determines the direction and strength of auxiliary alignment: large-drop pairs encourage sensitivity, reliable low-drop pairs receive bounded clean-anchored consistency, and ambiguous pairs are down-weighted or skipped. This design separates intervention construction from supervision assignment.

\item \textbf{SIVA-RL for scalable multimodal RL.}
We instantiate the formulation with token-aligned distance-constrained PatchSwap, frozen paired outcome valuation, clean-rollout re-scoring, and margin-bounded dual-direction alignment. SIVA-RL plugs into both GRPO and DAPO, consistently improves 3B and 7B backbones across nine benchmarks, and yields larger gains on vision-dependent reasoning. Ablations and training diagnostics validate the roles of localized intervention, soft routing, and sensitivity--invariance alignment.
\end{itemize}
\section{Related Work}
\label{sec:related_work}

\paragraph{Multimodal reinforcement learning with verifiable rewards.}
Reinforcement learning with verifiable rewards (RLVR) improves reasoning in
large language and vision-language models. GRPO enables critic-free
optimization through group-relative advantages. DAPO extends GRPO with
improved clipping, dynamic sampling, and entropy regularization for
large-scale training
\cite{shao2024deepseekmath,yu2026dapo}. Recent work brings RLVR to multimodal
reasoning through visual cold-start data, rule-verifiable rewards, and group-
or step-level policy optimization
\cite{huang2025vision,yao2025r1,meng2025mm,wang2026vl}. These methods share one
limitation: final-answer rewards give no direct signal about whether the
policy grounds its predictions in task-relevant visual evidence. SIVA-RL
complements these RLVR backbones. SIVA-RL augments policy optimization with
outcome-conditioned intervention alignment and instantiates with both GRPO-
and DAPO-style training.

\paragraph{Visual perception and intervention supervision.}
One line of work strengthens visual grounding through auxiliary perception
rewards, clean--intervention policy contrast, or token- and trajectory-level
visual credit assignment
\cite{xiao2025perception,wang2025perception,huang2025spotlight,
wang2026visually,li2026prpo}. Another line introduces distorted, masked,
mixed, or otherwise intervened inputs to improve visual exploration,
sensitivity, and robustness
\cite{liu2026noisyrollout,li2025vision,zhang2025cf,
gao2026thinking}. These methods establish the value of supervision beyond
answer correctness. Many of them, however, determine the optimization target
from the intervention type, or use interventions mainly to redistribute visual
credit across tokens. SIVA-RL instead evaluates the sample-wise behavioral
effect of each localized intervention. The paired outcome then determines both
the direction and strength of supervision. This design separates intervention
construction from outcome valuation and softly routes valid pairs toward
sensitivity alignment or bounded clean-anchored consistency.

\section{Method}
\label{sec:method}

\subsection{Overview and Base RL Backbone}
\label{sec:method_overview}

We consider a multimodal reasoning dataset
$\mathcal{D}=\{(q_i,x_i,a_i^\star)\}_{i=1}^{N}$, where $q_i$, $x_i$, and
$a_i^\star$ denote the question, image, and target answer, respectively.
For each clean input, the rollout policy samples a group of $G$ responses,
\begin{equation}
y_{i,g}
\sim
\pi_{\theta_{\mathrm{old}}}
\bigl(\cdot\mid q_i,x_i\bigr),
\qquad
g=1,\ldots,G.
\label{eq:clean_rollout}
\end{equation}
Each response receives a verifiable reward $r_{i,g}$; the backbone algorithm
then computes a group-relative advantage $A_{i,g}$ and the corresponding actor
update. We denote the full GRPO- or DAPO-style backbone objective as
$\mathcal{L}_{\mathrm{base}}$
\cite{shao2024deepseekmath,yu2026dapo}. Backbone-specific advantage
estimation, clipping, filtering, and regularization are detailed in the
appendix.

\begin{figure*}[t]
    \centering
    \includegraphics[width=0.98\textwidth]{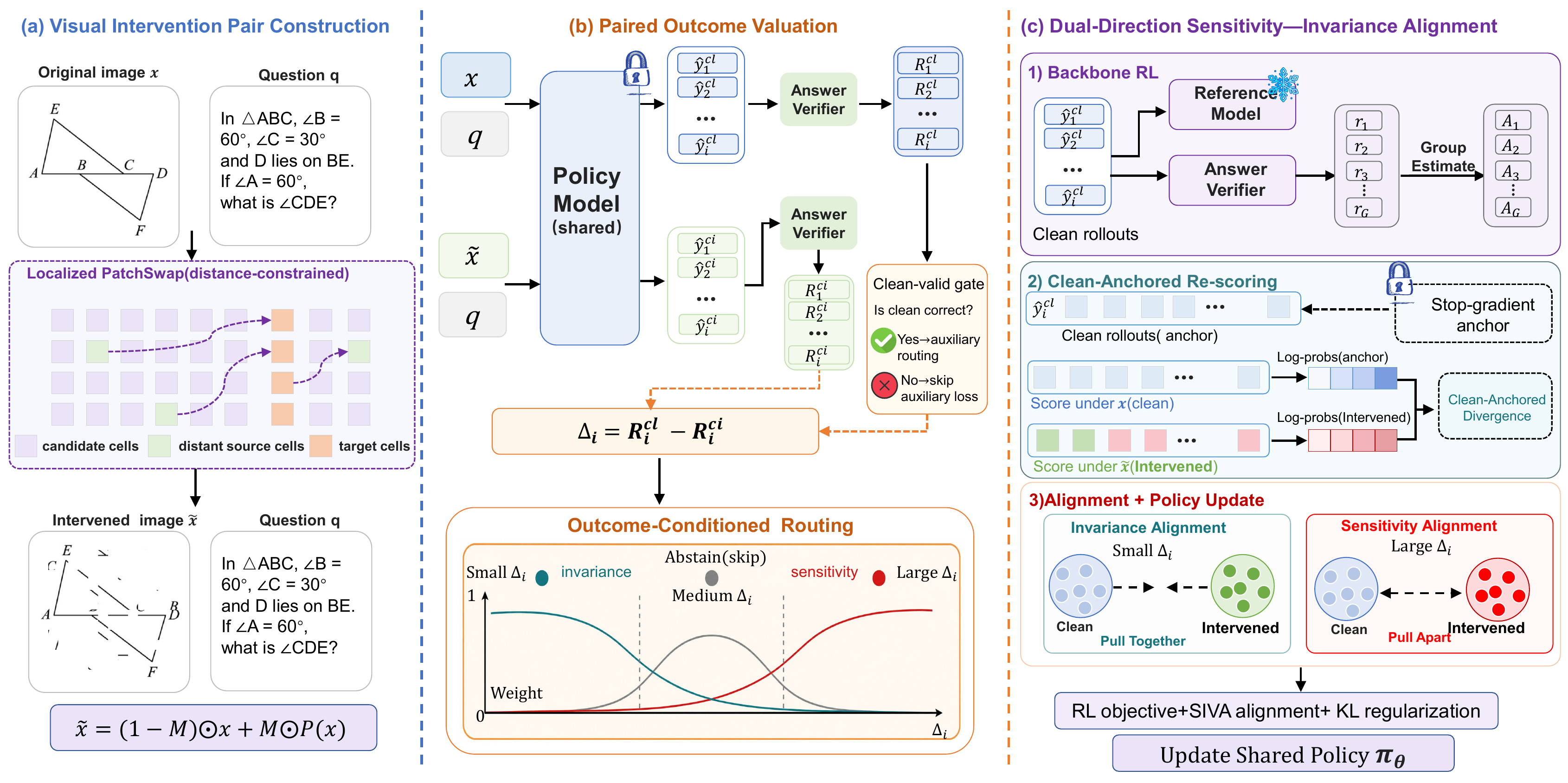}
    \caption{
    \textbf{Overview of SIVA-RL.}
    The upper branch represents the standard clean-image RLVR backbone.
    \textbf{(a)} Token-aligned, distance-constrained PatchSwap constructs a
    localized candidate intervention $\tilde{x}_i$.
    \textbf{(b)} A frozen audit policy performs paired decoding and
    converts the clean-to-intervention reward drop into sensitivity and
    invariance weights.
    \textbf{(c)} The clean rollout responses are re-scored under both visual
    conditions. Large-drop pairs are encouraged to exceed a sensitivity
    margin, while valid low-drop pairs are constrained within an invariance
    margin.
    }
    \label{fig:siva_method}
    \vspace{-10pt}
\end{figure*}

SIVA-RL preserves the clean-image backbone update and augments it with an
outcome-conditioned intervention-alignment objective. As illustrated in
Figure~\ref{fig:siva_method}, the framework has three components:
\textbf{(a)} localized intervention construction,
\textbf{(b)} outcome-conditioned soft routing, and
\textbf{(c)} clean-anchored dual-direction alignment. The routing branch
determines whether a candidate intervention reinforces sensitivity, encourages
invariance, or abstains. The alignment branch reuses existing clean-rollout
responses. This reuse avoids additional stochastic sampling beyond the paired
audit used for outcome valuation.

\subsection{Localized PatchSwap Construction}
\label{sec:intervention_construction}

Given an image $x_i$, we partition its pixel space into a grid
$\Omega_i$ aligned with the effective visual-token grid after patchification
and spatial merging. Let $p_{i,u}$ denote the image cell indexed by
$u\in\Omega_i$. We first sample candidate cells with ratio $\eta$ and then
select effective replacement targets with probability $\rho$:
\begin{gather}
  C_i \sim \operatorname{SampleMask}(\Omega_i,\eta), \quad C_{i,u}\in\{0,1\}, \\
  B_{i,u} \sim \operatorname{Bernoulli}(\rho), \\
  T_{i,u} = C_{i,u}B_{i,u}.
\end{gather}

Here $C_{i,u}=1$ marks cell $u$ as eligible for intervention; $T_{i,u}=1$
marks it as an active replacement target. Eligible but unselected cells appear
unmodified in the intervention.

For every target cell $u$, we form a valid within-image source set
\begin{equation}
\mathcal{S}_i(u)
=
\left\{
v\in\Omega_i:
C_{i,v}=0,\;
\bar d_i(u,v)\ge d_{\min}
\right\},
\qquad
s_i(u)\sim\operatorname{Unif}\bigl(\mathcal{S}_i(u)\bigr),
\label{eq:source_sampling}
\end{equation}
where $\bar d_i(u,v)\in[0,1]$ is the normalized grid distance between $u$ and
$v$, and $d_{\min}$ excludes spatially adjacent sources. The intervention view is then assembled as
\begin{equation}
\tilde p_{i,u}
=
\begin{cases}
p_{i,s_i(u)}, & T_{i,u}=1,\\
p_{i,u},      & T_{i,u}=0,
\end{cases}
\qquad
\tilde x_i
=
\operatorname{Assemble}
\bigl(\{\tilde p_{i,u}\}_{u\in\Omega_i}\bigr).
\label{eq:patchswap}
\end{equation}

Distance-constrained within-image PatchSwap disrupts localized evidence and
spatial relationships while preserving the image's global statistics and
style. The construction yields interventions harder than global masking yet
less destructive than complete visual removal. PatchSwap thus exposes
evidence-dependent reasoning without collapsing the visual context. Grid
resolution, candidate ratio $\eta$, replacement probability $\rho$, and
minimum distance $d_{\min}$ are specified in Appendix~\ref{sec:impl_details}.

\subsection{Outcome-Conditioned Soft Routing}
\label{sec:intervention_routing}

The same PatchSwap operator can produce different behavioral effects across
samples. SIVA-RL therefore assigns no fixed optimization role to constructed
interventions. SIVA-RL instead evaluates each clean--intervention pair with a
frozen audit policy $\pi_{\bar\theta}$, a no-gradient snapshot of the current
policy. For compactness, let
$x_i^{\mathrm{cl}}=x_i$ and $x_i^{\mathrm{ci}}=\tilde x_i$. Paired sampled
audit decoding and answer evaluation are defined by
\begin{equation}
\begin{aligned}
\hat y_{i,m}^{v} &\sim
\pi_{\bar\theta}(\cdot\mid q_i,x_i^v),\qquad
R_i^{v}=
\frac{1}{M}\sum_{m=1}^{M}
r_{\mathrm{ans}}(\hat y_{i,m}^{v},a_i^\star),
\quad v\in\{\mathrm{cl},\mathrm{ci}\},
\end{aligned}
\label{eq:paired_audit}
\end{equation}
where $r_{\mathrm{ans}}$ is the answer-level verifier used for
intervention valuation, and $R_i^v$ is the empirical verifier score over
the sampled audit rollouts for view $v$. The audit policy is fixed during the
update, and no gradient is propagated through the decoded answers or their
rewards.

We define the clean-anchored validity gate and observed reward drop as
\begin{equation}
\begin{aligned}
b_i&=\mathbb{I}[R_i^{\mathrm{cl}}\ge\tau_c]\,
\Phi_i^{\mathrm{cl}}\Phi_i^{\mathrm{ci}},
\qquad
\Delta_i=R_i^{\mathrm{cl}}-R_i^{\mathrm{ci}},
\end{aligned}
\label{eq:validity_drop}
\end{equation}
where $\Phi_i^v\in\{0,1\}$ indicates that the sampled audit outputs for view
$v$ contain a parseable response for the answer verifier. A
clean--intervention pair is \emph{valid} when $b_i=1$. The clean-outcome gate
prevents failed clean predictions from contributing spurious sensitivity or
invariance signals. To obtain routing weights, we define
\begin{equation}
\operatorname{Ramp}(z;a,b)=
\operatorname{clip}\!\left(\frac{z-a}{b-a},0,1\right),
\qquad a<b,
\label{eq:ramp}
\end{equation}
and compute
\begin{equation}
\begin{aligned}
w_i^{\mathrm{sen}}
&=b_i\,\operatorname{Ramp}
(\Delta_i;\delta_{\mathrm{sen}}^{\mathrm{lo}},
\delta_{\mathrm{sen}}^{\mathrm{hi}}),\\
w_i^{\mathrm{inv}}
&=b_i\left[1-\operatorname{Ramp}
(\Delta_i;\delta_{\mathrm{inv}}^{\mathrm{lo}},
\delta_{\mathrm{inv}}^{\mathrm{hi}})\right].
\end{aligned}
\label{eq:routing_weights}
\end{equation}
The sensitivity weight increases monotonically with the reward drop. The
invariance weight is highest for valid pairs where the intervention changes
little or nothing. When $\delta_{\mathrm{inv}}^{\mathrm{hi}} < \delta_{\mathrm{sen}}^{\mathrm{lo}}$,
the gap between the two thresholds defines an explicit abstention band where
neither branch receives weight. More generally, uncertain pairs receive
reduced auxiliary weights rather than hard intervention labels. The four
outcome transitions from Section~\ref{sec:introduction} serve as a diagnostic
lens only. Training operates on the empirical reward drop $\Delta_i$, the
validity gate $b_i$, and the resulting soft routing weights.

\subsection{Clean-Anchored Dual-Direction Alignment}
\label{sec:siva_alignment}

After routing, SIVA-RL reuses every response $y_{i,g}$ already sampled from the
clean-image rollout. The same token sequence is teacher-forced and re-scored
under the clean and intervention visual conditions using the trainable policy
$\pi_\theta$:
\begin{equation}
\ell_{i,g,t}^{v}
=\log\pi_\theta(y_{i,g,t}\mid q_i,x_i^v,y_{i,g,<t}),
\qquad v\in\{\mathrm{cl},\mathrm{ci}\}.
\label{eq:paired_rescoring}
\end{equation}
Because routing weights are defined at the input-pair level, all $G$
responses for example $i$ share the same $w_i^{\mathrm{sen}}$ and
$w_i^{\mathrm{inv}}$.

We treat the clean-conditioned likelihood as a stop-gradient anchor. Let
$\operatorname{sg}[\cdot]$ denote the stop-gradient operator. The
clean-anchored token discrepancy and its response-level average are
\begin{equation}
\begin{gathered}
z_{i,g,t}=\ell_{i,g,t}^{\mathrm{cf}}
-\operatorname{sg}[\ell_{i,g,t}^{\mathrm{cl}}],
\qquad
d_{i,g,t}=\exp(z_{i,g,t})-z_{i,g,t}-1,\\
D_{i,g}=\frac{1}{|\mathcal{T}_{i,g}|}
\sum_{t\in\mathcal{T}_{i,g}}d_{i,g,t},
\end{gathered}
\label{eq:clean_anchored_divergence}
\end{equation}
where $\mathcal{T}_{i,g}$ denotes the valid response-token set. The quantity
$d_{i,g,t}$ is a non-negative Monte Carlo surrogate for the token-level
clean-to-intervention KL divergence. Only tokens drawn from the clean rollout
are evaluated, so $D_{i,g}$ approximates the full-vocabulary KL rather than
computing it exactly. SIVA-RL applies two margin-based objectives:
\begin{equation}
\begin{aligned}
\mathcal{L}_{\mathrm{sen}}
&=
\frac{\sum_{i,g}w_i^{\mathrm{sen}}[m_{\mathrm{sen}}-D_{i,g}]_+}
{\sum_{i,g}w_i^{\mathrm{sen}}+\epsilon},\\
\mathcal{L}_{\mathrm{inv}}
&=
\frac{\sum_{i,g}w_i^{\mathrm{inv}}[D_{i,g}-m_{\mathrm{inv}}]_+}
{\sum_{i,g}w_i^{\mathrm{inv}}+\epsilon},
\end{aligned}
\label{eq:dual_alignment}
\end{equation}
where $[z]_+=\max(z,0)$. For a reliable high-drop pair,
$\mathcal{L}_{\mathrm{sen}}$ enforces a minimum divergence of
$m_{\mathrm{sen}}$ between clean and intervention policies. For a valid
low-drop pair, $\mathcal{L}_{\mathrm{inv}}$ constrains this discrepancy below
$m_{\mathrm{inv}}$. Thus, \emph{dual-direction} refers to pushing apart versus
pulling together under the same clean-anchored measure---not to a symmetric
divergence objective. The complete objective is
\begin{equation}
\mathcal{L}_{\mathrm{SIVA}}
=\mathcal{L}_{\mathrm{base}}
+\lambda_{\mathrm{sen}}\mathcal{L}_{\mathrm{sen}}
+\lambda_{\mathrm{inv}}\mathcal{L}_{\mathrm{inv}},
\label{eq:full_objective}
\end{equation}
where $\lambda_{\mathrm{sen}}$ and $\lambda_{\mathrm{inv}}$ control the two
auxiliary branches. Setting either to zero recovers the corresponding
single-branch variant. All backbone-specific regularizers are absorbed into
$\mathcal{L}_{\mathrm{base}}$; routing thresholds, margins, and loss
coefficients are reported in Appendix~\ref{sec:impl_details}.

\section{Experiments}
\label{sec:experiments}

\definecolor{genblue}{RGB}{220,240,248}
\definecolor{visorange}{RGB}{252,232,215}
\definecolor{overallpurple}{RGB}{238,221,240}
\definecolor{relgreen}{RGB}{120,220,120}
\definecolor{avggray}{RGB}{247,247,247}

\newcommand{\relup}[1]{\cellcolor{relgreen}{$\uparrow$ #1}}
\newcommand{\gain}[2]{#1{\scriptsize\textcolor{green!60!black}{\,(+#2)}}}
\newcommand{\gainb}[2]{\textbf{#1}{\scriptsize\textcolor{green!60!black}{\,(+#2)}}}
\newcommand{\gainu}[2]{\underline{#1}{\scriptsize\textcolor{green!60!black}{\,(+#2)}}}
\newcommand{\gainds}[2]{#1{\scriptsize\textcolor{green!60!black}{\,(#2)}}}
\newcommand{\gaindbs}[2]{\textbf{#1}{\scriptsize\textcolor{green!60!black}{\,(#2)}}}
\newcommand{\gaindus}[2]{\underline{#1}{\scriptsize\textcolor{green!60!black}{\,(#2)}}}

We evaluate SIVA-RL across multiple multimodal RL backbones and model scales.
The main evaluation covers nine benchmarks spanning mathematical, logical,
counting, and multi-disciplinary multimodal reasoning. We also compare SIVA-RL
with reported results from proprietary, general-purpose, and math-oriented
multimodal VLMs. Appendix~\ref{sec:appendix_medical} provides additional
medical-domain evaluations. The experiments answer four questions:
\textbf{(1)} Does SIVA-RL consistently improve its corresponding RL backbone
across model scales?
\textbf{(2)} How competitive are the resulting models under broader multimodal reasoning comparisons?
\textbf{(3)} How do intervention construction and outcome-conditioned
alignment contribute to the final performance?
\textbf{(4)} Does the same PatchSwap operator continue to induce heterogeneous routing behavior during training?

\subsection{Experimental Setup}
\label{sec:experimental_setup}
\textbf{Models and RL backbones.}
We experiment with Qwen2.5-VL-3B and Qwen2.5-VL-7B \cite{bai2025qwen25vl}
on both GRPO- and DAPO-style policy-optimization backbones
\cite{shao2024deepseekmath,yu2026dapo}, training on ViRL39K
\cite{wang2026vl}. All backbone--scale configurations share the same training
data and SIVA-RL hyperparameters. For each configuration, SIVA-RL augments the
corresponding RL objective with localized intervention construction,
outcome-conditioned routing, and dual-direction alignment
(Section~\ref{sec:method}). Training applies a uniform-sensitivity warm-up
followed by outcome-conditioned dual-direction alignment.
Appendix~\ref{sec:impl_details} provides the optimization and validation
settings.\\
\textbf{Benchmarks and metrics.} Following prior perception-aware multimodal RL evaluation protocols
\cite{wang2025perception,wang2026visually}, we organize the main benchmark suite into
two groups. The general multimodal reasoning group includes Geometry3K
(Geo3K) \cite{lu2021inter}, MathVista \cite{lu2024mathvista},
We-Math \cite{qiao2025wemath}, MMK12 \cite{meng2025mm}, and
MathVerse \cite{zhang2024mathverse}. The vision-dependent group includes
LogicVista \cite{xiao2024logicvista}, the counting subset of Super-CLEVR
\cite{li2023super}, MMMU-Pro \cite{yue2025mmmu}, and the
vision-only subset of MathVerse, denoted MathVerse-V
\cite{zhang2024mathverse}. We report Avg@8 accuracy (\%) for each benchmark.
The general, vision-dependent, and overall scores are benchmark-level macro
averages.\\

\textbf{External results.}
Results for external models are taken from their original papers or official
reports, and unavailable entries are denoted by ``--''. Benchmark coverage
differs across models, so averages computed from available entries serve as
references rather than a strict unified-protocol leaderboard.
Appendix~\ref{sec:impl_details} provides the implementation and evaluation
details.

\begin{table*}[t!]
\centering
\caption{\textbf{Main results on general multimodal reasoning, vision-dependent multimodal reasoning, and overall performance with Qwen2.5-VL-3B and Qwen2.5-VL-7B as backbones.} $\Delta_{\mathrm{rel}}^{\%}$ denotes the relative overall improvement over the matched RL baseline at the same scale.}
\label{tab:reasoning_results}
\vspace{-8pt}
\footnotesize
\setlength{\tabcolsep}{1.6pt}
\renewcommand{\arraystretch}{1.1}
\resizebox{\textwidth}{!}{
\begin{tabular}{l|cccccc|ccccc|cc}
\toprule
\multirow{2}{*}{\textbf{Method}}
& \multicolumn{6}{c|}{\cellcolor{genblue}\textbf{General Multimodal Reasoning}}
& \multicolumn{5}{c|}{\cellcolor{visorange}\textbf{Vision-Dependent Multimodal Reasoning}}
& \multicolumn{2}{c}{\cellcolor{overallpurple}\textbf{Overall}} \\
\cmidrule(lr){2-7} \cmidrule(lr){8-12} \cmidrule(lr){13-14}
& \textbf{Geo3k} & \textbf{MathVista} & \textbf{We-Math} & \textbf{MMK12} & \textbf{MathVerse} & \textbf{AVG}
& \textbf{LogicVista} & \textbf{Counting} & \textbf{MMMU-Pro} & \textbf{MathVerse$_V$} & \textbf{AVG}
& \textbf{AVG} & $\Delta_{\mathrm{rel}}^{\%}$ \\
\midrule
GRPO-3B & 28.72 & 59.34 & 58.90 & 57.24 & 55.25 & 51.89 & 38.14 & 55.81 & 25.66 & 52.26 & 42.97 & 47.92 & -- \\
\textbf{GRPO-3B + SIVA-RL} & 34.26 & 62.12 & 61.92 & 64.63 & 59.66 & 56.52 & 42.42 & 75.19 & 29.12 & 55.68 & 50.60 & 53.89 & \textcolor{green!60!black}{+12.46} \\
\midrule
GRPO-7B & 40.18 & 65.48 & 68.12 & 72.26 & 66.51 & 62.51 & 45.62 & 73.94 & 35.17 & 61.71 & 54.11 & 58.78 & -- \\
\textbf{GRPO-7B + SIVA-RL} & 44.41 & 69.24 & 69.53 & 76.03 & 69.29 & 65.70 & 45.83 & 88.75 & 37.43 & 65.64 & 59.42 & 62.90 & \textcolor{green!60!black}{+7.01} \\
\midrule
DAPO-3B & 31.20 & 60.89 & 59.95 & 66.83 & 56.25 & 55.02 & 40.69 & 74.25 & 28.42 & 53.09 & 49.11 & 52.40 & -- \\
\textbf{DAPO-3B + SIVA-RL} & 37.81 & 61.41 & 62.59 & 68.74 & 61.18 & 58.35 & 43.34 & 80.25 & 30.04 & 58.49 & 53.03 & 55.98 & \textcolor{green!60!black}{+6.83} \\
\midrule
DAPO-7B & 35.92 & 61.91 & 58.51 & 75.93 & 55.64 & 57.58 & 37.05 & 90.05 & 29.02 & 51.04 & 51.79 & 55.01 & -- \\
\textbf{DAPO-7B + SIVA-RL} & 42.95 & 64.09 & 70.50 & 79.68 & 69.73 & 65.39 & 47.23 & 92.25 & 36.94 & 65.91 & 60.58 & 63.25 & \textcolor{green!60!black}{+14.98} \\
\bottomrule
\end{tabular}}
\vspace{-8pt}
\end{table*}

\begin{table*}[t!]
    \centering
    \caption{
    \textbf{Comparison with reported results of proprietary, general-purpose, and
    math-oriented multimodal VLMs.} Bold and underlined denote the best and second-best results; green subscripts indicate gains over the base model.}

    \label{tab:math_reasoning}
    \vspace{-8pt}

    \small
    \setlength{\tabcolsep}{4pt}
    \renewcommand{\arraystretch}{1.08}

    \resizebox{\textwidth}{!}{
    \begin{tabular}{llllllll}
        \toprule
        \multirow{2}{*}{\textbf{Method}}
        & \multicolumn{7}{c}{
            \cellcolor{gray!20}
            \textbf{Mathematical Multimodal Reasoning}
        } \\
        \cmidrule(lr){2-8}
        & \textbf{Geo3k}
        & \textbf{MathVista}
        & \textbf{We-Math}
        & \textbf{MathVerse}
        & \textbf{MathVerse$_V$}
        & \textbf{MMK12}
        & \textbf{AVG} \\
        \midrule

        \multicolumn{8}{c}{\textbf{Proprietary Models}} \\
        \midrule

        GPT-4o \cite{openai2024gpt4o}
        & -- & 64.7 & 62.8 & 50.2 & 53.8 & 55.8 & 57.5 \\

        GPT-4o-mini \cite{openai2024gpt4omini}
        & -- & 59.9 & 56.3 & 42.3 & 45.1 & 51.9 & 51.1 \\

        Gemini-2.0-Flash \cite{google2024gemini20}
        & -- & 70.4 & 47.4 & 47.8 & 48.7 & 65.2 & 55.9 \\

        \midrule
        \multicolumn{8}{c}{
            \textbf{General-Purpose Multimodal VLMs}
        } \\
        \midrule

        Qwen2.5-VL-72B \cite{bai2025qwen25vl}
        & -- & \textbf{74.2} & 49.1 & 47.3 & 48.6 & 70.5 & 57.9 \\

        InternVL2.5-8B \cite{chen2024expanding}
        & -- & 64.9 & 44.9 & 37.0 & 40.2 & 46.8 & 46.8 \\

        InternVL2.5-78B \cite{chen2024expanding}
        & -- & 64.9 & 44.9 & 37.0 & 40.2 & 59.8 & 49.4 \\

        LLaVA-OneVision-7B \cite{li2024llava}
        & -- & 58.5 & 44.1 & -- & -- & -- & 51.3 \\

        LLaVA-OneVision-72B \cite{li2024llava}
        & -- & 67.1 & 32.0 & 27.2 & 30.1 & -- & 39.1 \\

        LLaVA-OneVision-1.5-8B \cite{an2025llava}
        & -- & 69.6 & 61.5 & -- & -- & -- & 65.6 \\

        LLaVA-Critic-R1-7B \cite{wang2025llava}
        & 35.4 & 68.7 & 62.6 & 58.9 & 53.1 & 57.4 & 56.0 \\

        R1-OneVision-7B \cite{yang2025r1}
        & 30.6 & 64.9 & 55.2 & 61.7 & 44.3 & 43.3 & 50.0 \\

        \midrule
        \multicolumn{8}{c}{
            \textbf{Math-Specific Multimodal VLMs}
        } \\
        \midrule

        MM-Eureka-7B \cite{meng2025mm}
        & 36.4 & 59.1 & 45.3 & 57.6 & 56.4 & 60.6 & 52.6 \\

        MM-Eureka-7B-CPGD
        \cite{meng2025mm}
        & 37.6 & 64.2 & 64.3 & 63.7 & 59.2 & 64.7 & 59.0 \\

        ADORA-7B \cite{gui2025training}
        & 41.2 & 61.1 & 53.0 & 45.2 & 41.8 & 49.8 & 48.7 \\

        R1-VL-7B \cite{zhang2025r1vl} 
        & 31.9 & 63.5 & 56.1 & 42.0 & 43.2 & 55.3 & 48.7 \\

        VLAA-Thinker-7B \cite{chen2025sft}
        & 24.2 & 67.4 & 65.9 & 47.9 & 52.0 & 63.2 & 53.4 \\

        VL-Rethinker-7B \cite{wang2026vl}
        & 33.6 & 61.3 & 66.5 & 64.0 & 60.8 & 59.8 & 57.7 \\

        RACRO-7B-CRO-GRPO \cite{gou2025reasoning}
        & 41.4 & 61.7 & 68.9 & 65.7 & 61.7 & 70.5 & 61.7 \\

        ThinkLite-VL-7B \cite{wang2026sota}
        & 34.4 & 68.9 & 63.5 & 49.5 & 46.0 & 56.2 & 53.1 \\
        \midrule
        \multicolumn{8}{c}{
            \textbf{RL-based Qwen2.5-VL Variants}
        } \\
        \midrule

        PAPO-3B~\cite{wang2025perception}
        & 31.0 & 61.4 & 60.1 & 57.1 & 54.0 & 57.4 & 53.5 \\

        VPPO-3B~\cite{huang2025spotlight}
        & 35.8 & 63.6 & 61.4 & 61.3 & 58.1 & 65.3 & 57.6 \\

        DVRP-3B~\cite{gao2026thinking}
        & 34.5 & 65.5 & 60.3 & 57.7 & 54.5 & 61.2 & 55.6 \\

        PAPO-7B~\cite{wang2025perception}
        & 40.3 & 69.5 & 66.8 & 68.4 & 65.0 & 72.5 & 63.7 \\

        DVRP-7B~\cite{gao2026thinking}
        & 43.4 & 70.9 & 67.8 & 68.9 & 65.3 & 74.1 & 65.1 \\

        \midrule
        \multicolumn{8}{c}{\textbf{Our Models}} \\
        \midrule

        Qwen2.5-VL-3B (base) 
        & 20.6 & 40.6 & 23.9 & 30.9 & 28.2 & 34.8 & 29.8 \\

        \hspace{2em}DAPO + SIVA-RL 
        & \gain{37.8}{17.2}
        & \gain{61.4}{20.8}
        & \gain{62.6}{38.7}
        & \gain{61.2}{30.3}
        & \gain{58.5}{30.3}
        & \gain{68.7}{33.9}
        & \gain{58.4}{28.6} \\

        \hspace{2em}GRPO + SIVA-RL
        & \gain{34.3}{13.7}
        & \gain{62.1}{21.5}
        & \gain{61.9}{38.0}
        & \gain{59.7}{28.8}
        & \gain{55.7}{27.5}
        & \gain{64.6}{29.8}
        & \gain{56.4}{26.6} \\

        Qwen2.5-VL-7B (base) 
        & 33.8 & 55.9 & 41.8 & 45.6 & 36.9 & 43.7 & 43.0 \\

        \hspace{2em}DAPO + SIVA-RL 
        & \gainu{43.0}{9.2}
        & \gain{64.1}{8.2}
        & \gainb{70.5}{28.7}
        & \gainb{69.7}{24.1}
        & \gainb{65.9}{29.0}
        & \gainb{79.7}{36.0}
        & \gainu{65.5}{22.5} \\

        \hspace{2em}GRPO + SIVA-RL 
        & \gainb{44.4}{10.6}
        & \gain{69.2}{13.3}
        & \gainu{69.5}{27.7}
        & \gainu{69.3}{23.7}
        & \gainu{65.6}{28.7}
        & \gainu{76.0}{32.3}
        & \gainb{65.7}{22.7} \\

        \bottomrule
    \end{tabular}
    }

    \vspace{-14pt}
\end{table*}

\subsection{Main Results}
\label{sec:main_results}

\paragraph{Improvements over matched RL backbones.}
SIVA-RL consistently improves all four matched backbone--scale configurations
in Table~\ref{tab:reasoning_results}. Absolute overall gains range from
$3.58$ to $8.24$ points, corresponding to relative improvements from
$6.83\%$ to $14.98\%$ over the matched RL baseline. The gains hold for both
GRPO and DAPO, and for both 3B and 7B models, indicating that SIVA-RL is not
tied to a specific optimizer or model scale.

The improvements are larger on the vision-dependent subset in every matched
comparison. For instance, GRPO-3B gains $+7.63$ points on vision-dependent
benchmarks compared with $+4.63$ points on general benchmarks, and DAPO-7B
shows a similar pattern with $+8.79$ versus $+7.81$ points. This suggests that
SIVA-RL provides more than a generic RL enhancement: it is particularly
effective when correct reasoning depends on visual evidence, which is
consistent with the goal of outcome-conditioned sensitivity--invariance
alignment.

\paragraph{Comparison with existing multimodal VLMs.}
The 7B SIVA-RL models achieve the strongest full-suite averages among the
reported 7B systems and perform strongly on We-Math, MathVerse, MathVerse-V,
and MMK12 (Table~\ref{tab:math_reasoning}). The 3B variants remain competitive
with several much larger models. Some external results cover only subsets of
the six benchmarks, so their partial averages serve as context rather than a
unified leaderboard.

\subsection{Ablation Studies}
\label{sec:ablations}

\paragraph{Intervention Construction.}
\label{sec:construction_ablation}

\begin{table*}[t]
\centering
\caption{\textbf{Ablation of intervention construction and outcome-conditioned alignment on Qwen2.5-VL-3B with GRPO.}
PatchSwap w/o dist. sets $d_{\min}=0.00$; PatchSwap denotes the default distance-constrained construction.
$\Delta_{\mathrm{base}}$ denotes the relative change of Overall AVG over the GRPO-3B baseline.}
\label{tab:construction_ablation}

\scriptsize
\setlength{\tabcolsep}{2.6pt}
\renewcommand{\arraystretch}{1.04}
\newcommand{\posd}[1]{\textcolor{green!50!black}{#1}}

\resizebox{\textwidth}{!}{
\begin{tabular}{l|ccccc|cccc|cc}
\toprule
\multirow{2}{*}{\textbf{Setting}}
& \multicolumn{5}{c|}{\cellcolor{genblue}\textbf{General Multimodal Reasoning}}
& \multicolumn{4}{c|}{\cellcolor{visorange}\textbf{Vision-Dependent Multimodal Reasoning}}
& \multicolumn{2}{c}{\cellcolor{overallpurple}\textbf{Overall}} \\
\cmidrule(lr){2-6} \cmidrule(lr){7-10} \cmidrule(lr){11-12}
& \textbf{Geo3k} & \textbf{MathVista} & \textbf{We-Math} & \textbf{MMK12} & \textbf{MathVerse}
& \textbf{LogicVista} & \textbf{Counting} & \textbf{MMMU-Pro} & \textbf{MathVerse$_V$}
& \textbf{AVG} & $\boldsymbol{\Delta}_{\mathrm{base}}$ \\
\midrule
GRPO-3B baseline
& 28.72 & 59.34 & 58.90 & 57.24 & 55.25
& 38.14 & 55.81 & 25.66 & 52.26
& 47.92 & -- \\

Random Mask
& 30.95 & 61.38 & 60.09 & 57.39 & 57.14
& 38.67 & 62.56 & 27.11 & 53.95
& 49.92 & \posd{+4.17\%} \\

Random Mask + SIVA-RL
& 31.47 & 60.40 & 60.01 & 59.42 & 58.20
& 38.93 & 68.62 & 28.11 & 54.95
& 51.12 & \posd{+6.68\%} \\

PatchSwap w/o dist.
& 30.30 & 58.02 & 59.60 & 52.01 & 56.39
& 38.93 & 63.38 & 26.08 & 52.99
& 48.64 & \posd{+1.50\%} \\

PatchSwap w/o dist. + SIVA-RL
& 33.96 & \textbf{62.64} & 61.28 & 60.38 & 59.00
& 42.20 & \textbf{75.38} & 28.30 & 55.66
& 53.20 & \posd{+11.02\%} \\

PatchSwap
& 31.66 & 59.36 & 60.89 & 55.61 & 58.14
& 40.35 & 64.50 & 27.63 & 55.13
& 50.36 & \posd{+5.09\%} \\

\textbf{PatchSwap + SIVA-RL}
& \textbf{34.26} & 62.12 & \textbf{61.92} & \textbf{64.63} & \textbf{59.66}
& \textbf{42.42} & 75.19 & \textbf{29.12} & \textbf{55.68}
& \textbf{53.89} & \posd{\textbf{+12.46\%}} \\
\bottomrule
\end{tabular}}
\vspace{-6pt}
\end{table*}

Table~\ref{tab:construction_ablation} disentangles the effects of intervention
construction and outcome-conditioned alignment. Among the construction-only
variants, distance-constrained PatchSwap achieves the highest overall average
($50.36$). This variant outperforms random masking ($49.92$) and unconstrained
PatchSwap ($48.64$). A minimum source distance avoids overly local
replacements and preserves more visual context than random masking.

Outcome-conditioned routing improves all three operators, reaching $51.12$,
$53.20$, and $53.89$ with random masking, unconstrained PatchSwap, and
distance-constrained PatchSwap. The best result combines SIVA-RL with
distance-constrained PatchSwap. This combination demonstrates the
complementary roles of intervention construction and sample-wise outcome
valuation.

\paragraph{Outcome-Conditioned Alignment.}
\label{sec:alignment_ablation}

\begin{table*}[t]
\centering
\caption{\textbf{Stepwise ablation of outcome-conditioned alignment on Qwen2.5-VL-3B with a GRPO backbone.}
Complete results are reported on all nine benchmarks.
$\Delta_{\mathrm{step}}$ is the relative change of Overall AVG over the previous row, showing the incremental contribution of each step.}
\label{tab:alignment_ablation}

\vspace{-8pt}
\scriptsize
\setlength{\tabcolsep}{2.6pt}
\renewcommand{\arraystretch}{1.05}
\newcommand{\posstep}[1]{\textcolor{green!50!black}{#1}}

\resizebox{\textwidth}{!}{
\begin{tabular}{l|ccccc|cccc|cc}
\toprule
\multirow{2}{*}{\textbf{Setting}}
& \multicolumn{5}{c|}{\cellcolor{genblue}\textbf{General Multimodal Reasoning}}
& \multicolumn{4}{c|}{\cellcolor{visorange}\textbf{Vision-Dependent Multimodal Reasoning}}
& \multicolumn{2}{c}{\cellcolor{overallpurple}\textbf{Overall}} \\
\cmidrule(lr){2-6} \cmidrule(lr){7-10} \cmidrule(lr){11-12}
& \textbf{Geo3k} & \textbf{MathVista} & \textbf{We-Math} & \textbf{MMK12} & \textbf{MathVerse}
& \textbf{LogicVista} & \textbf{Counting} & \textbf{MMMU-Pro} & \textbf{MathVerse$_V$}
& \textbf{AVG} & $\boldsymbol{\Delta}_{\mathrm{step}}$ \\
\midrule
GRPO-3B baseline
& 28.72 & 59.34 & 58.90 & 57.24 & 55.25
& 38.14 & 55.81 & 25.66 & 52.26
& 47.92 & -- \\

+PatchSwap
& 31.66 & 59.36 & 60.89 & 55.61 & 58.14
& 40.35 & 64.50 & 27.63 & 55.13
& 50.36 & \posstep{+5.09\%} \\

+Hard Sens.
& 31.32 & 60.74 & 61.49 & 61.64 & 57.82
& 42.00 & 65.06 & 27.67 & 54.82
& 51.40 & \posstep{+2.07\%} \\

+Soft Sens.
& 32.30 & \textbf{63.90} & 59.15 & 62.58 & 56.46
& 39.49 & 71.56 & 27.64 & 53.31
& 51.82 & \posstep{+2.90\%} \\

+Soft Sens. + Warmup (w/o Inv.)
& \textbf{34.48} & 61.36 & \textbf{62.68} & 61.88 & 58.37
& 40.75 & 71.81 & 27.45 & 55.00
& 52.64 & \posstep{+1.58\%} \\

+\textbf{SIVA-RL}
& 34.26 & 62.12 & 61.92 & \textbf{64.63} & \textbf{59.66}
& \textbf{42.42} & \textbf{75.19} & \textbf{29.12} & \textbf{55.68}
& \textbf{53.89} & \posstep{\textbf{+2.37\%}} \\
\bottomrule
\end{tabular}}
\vspace{-15pt}
\end{table*}

Table~\ref{tab:alignment_ablation} evaluates the alignment components while
holding distance-constrained PatchSwap fixed. Hard and soft sensitivity are
alternative routing strategies. Relative to PatchSwap alone ($50.36$), they
increase the overall average to $51.40$ and $51.82$. The gain from soft
sensitivity shows that outcome changes should not be treated as a binary
signal. The magnitude of the clean-to-intervention reward drop instead
provides a smoother and more reliable supervision strength. The advantage is
clearest when interventions induce partial evidence corruption rather than
complete answer flips.

The uniform-sensitivity warm-up raises performance to $52.64$. Early-stage
optimization therefore benefits from a stabilized alignment signal before
outcome-conditioned routing. Invariance alignment for clean-anchored low-drop
pairs yields the complete SIVA-RL objective and raises the overall average to
$53.89$. This step adds a $1.85$-point gain on the vision-dependent subset.
Low-drop interventions are therefore not uninformative; they provide useful
clean-anchored consistency constraints. These results support the
complementary roles of stabilized sensitivity learning and conservative
alignment for valid low-drop interventions.

\begin{figure*}[!t]
    \centering
    \includegraphics[width=\textwidth]{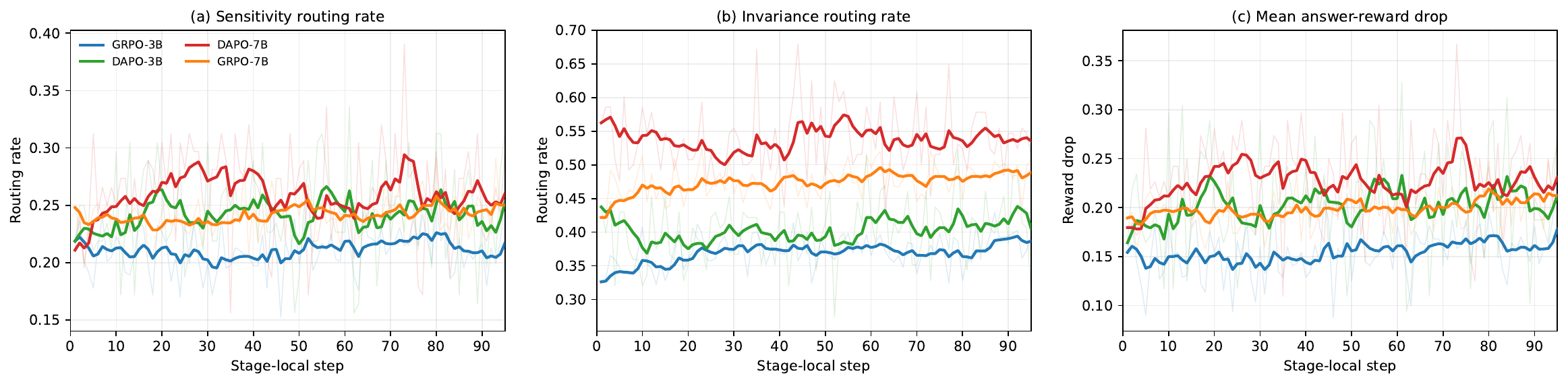}
    \caption{
    \textbf{Training-time outcome-conditioned routing diagnostics.}
    We report the sensitivity routing rate, invariance routing rate, and mean
    paired answer-reward drop for the four backbone--scale configurations.
    }
    \label{fig:neg_posi_drop}
    \vspace{-16pt}
\end{figure*}

\paragraph{Hyperparameter Sensitivity.}
\label{sec:hparam_sensitivity}

\begin{wrapfigure}[15]{r}{0.45\textwidth}
    \centering
    \vspace{-4pt}
    \includegraphics[
        width=0.98\linewidth
    ]{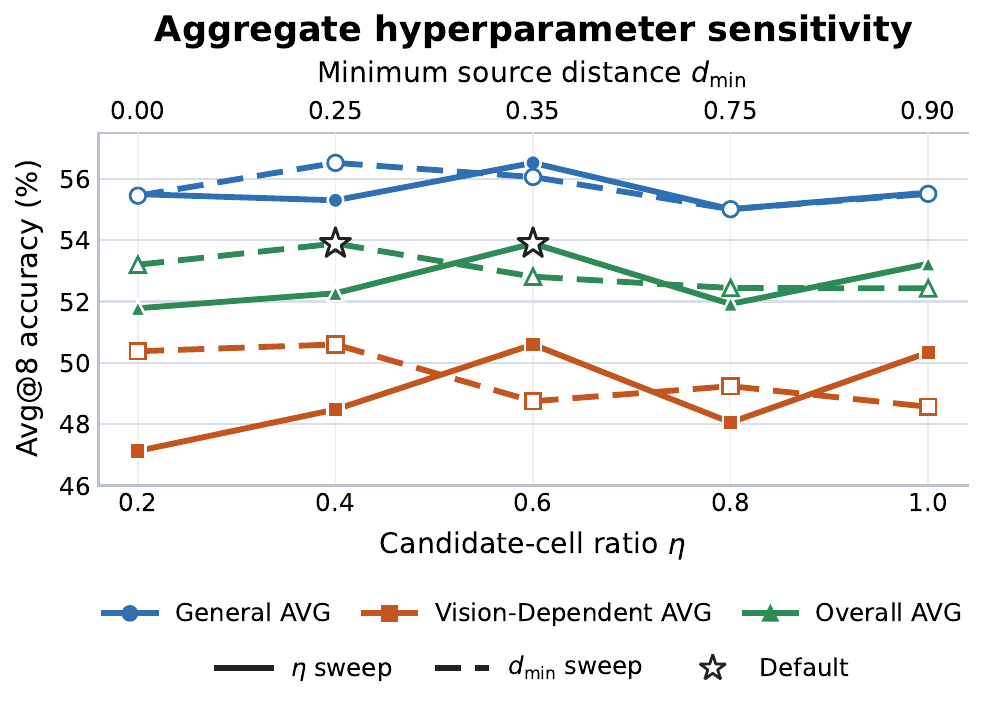}
    \vspace{-6pt}
    \caption{
    \textbf{Hyperparameter sensitivity.}
    Macro Avg@8 under varying $\eta$ and $d_{\min}$.
    }
    \label{fig:hparam_aggregate_sensitivity}
    \vspace{-4pt}
\end{wrapfigure}

Figure~\ref{fig:hparam_aggregate_sensitivity} studies the strength and spatial
extent of the PatchSwap intervention. The default setting $\eta=0.6$ and
$d_{\min}=0.25$ yields the strongest aggregate performance across the reported
averages. A moderate candidate-cell ratio balances preservation of the original visual context against modification of sufficient local evidence.
Overly weak or overly dense interventions reduce performance on part of the
benchmark suite. A moderate source-distance constraint is likewise preferable to either unconstrained replacement or overly restrictive source selection.

The $\eta=1.0$ configuration is a stress test. When all cells are eligible
candidates and no standard non-candidate source remains, the implementation
invokes the fallback source-sampling rule described in
Appendix~\ref{sec:impl_details}. The $\eta=1.0$ setting therefore probes the
robustness of the construction procedure under an extreme intervention ratio,
not the default source-selection regime.

\subsection{Mechanistic Analysis}
\label{sec:mechanistic_analysis}

We examine whether SIVA-RL maintains heterogeneous supervision after the
200-step warm-up. We track the paired answer-reward drop and the fractions of
pairs receiving non-zero sensitivity or invariance weights.
Figure~\ref{fig:neg_posi_drop} shows that both branches remain active across
all four backbone--scale configurations. Sensitivity routing increases with
the observed reward drop, while invariance routing stays substantial
throughout training. The same PatchSwap operator therefore yields both
answer-changing and low-drop pairs. Importantly, neither branch collapses to a
single dominant regime: sensitivity remains non-negligible for visually fragile
pairs, whereas invariance is consistently assigned to a large fraction of
low-drop pairs. The mean reward drop also stays in a moderate range rather than
vanishing, suggesting that the model does not simply learn to ignore the
intervention. This behavior supports the sample-wise dual-direction routing
underlying SIVA-RL.

\section{Conclusion}

We introduced SIVA-RL, an outcome-conditioned visual alignment framework
motivated by the heterogeneous sample-level effects of visual interventions.
SIVA-RL separates intervention construction from supervision assignment. It
routes each clean--intervention pair toward sensitivity or clean-anchored
invariance alignment according to the observed reward change. Across GRPO- and
DAPO-based 3B and 7B models, SIVA-RL improves performance on nine multimodal
reasoning benchmarks, with the strongest gains on vision-dependent tasks.
Ablations and training-time diagnostics validate the complementary roles of
localized PatchSwap, soft outcome-conditioned routing, and dual-direction
alignment. These results establish sample-wise outcome valuation as an
effective alternative to operator-defined intervention supervision in
multimodal RL.

\bibliography{iclr2025_conference}

@article{shao2024deepseekmath,
  title={Deepseekmath: Pushing the limits of mathematical reasoning in open language models},
  author={Shao, Zhihong and Wang, Peiyi and Zhu, Qihao and Xu, Runxin and Song, Junxiao and Bi, Xiao and Zhang, Haowei and Zhang, Mingchuan and Li, YK and Wu, Yang and others},
  journal={arXiv preprint arXiv:2402.03300},
  year={2024}
}

@article{guo2025deepseek,
  title={Deepseek-r1: Incentivizing reasoning capability in llms via reinforcement learning},
  author={Guo, Daya and Yang, Dejian and Zhang, Haowei and Song, Junxiao and Wang, Peiyi and Zhu, Qihao and Xu, Runxin and Zhang, Ruoyu and Ma, Shirong and Bi, Xiao and others},
  journal={arXiv preprint arXiv:2501.12948},
  year={2025}
}

@article{huang2025vision,
  title={Vision-r1: Incentivizing reasoning capability in multimodal large language models},
  author={Huang, Wenxuan and Jia, Bohan and Zhai, Zijie and Cao, Shaosheng and Ye, Zheyu and Zhao, Fei and Xu, Zhe and Tang, Xu and Hu, Yao and Lin, Shaohui},
  journal={arXiv preprint arXiv:2503.06749},
  year={2025}
}

@article{yao2025r1,
  title={R1-sharevl: Incentivizing reasoning capability of multimodal large language models via share-grpo},
  author={Yao, Huanjin and Yin, Qixiang and Zhang, Jingyi and Yang, Min and Wang, Yibo and Wu, Wenhao and Su, Fei and Shen, Li and Qiu, Minghui and Tao, Dacheng and others},
  journal={arXiv preprint arXiv:2505.16673},
  year={2025}
}

@article{meng2025mm,
  title={Mm-eureka: Exploring the frontiers of multimodal reasoning with rule-based reinforcement learning},
  author={Meng, Fanqing and Du, Lingxiao and Liu, Zongkai and Zhou, Zhixiang and Lu, Quanfeng and Fu, Daocheng and Han, Tiancheng and Shi, Botian and Wang, Wenhai and He, Junjun and others},
  journal={arXiv preprint arXiv:2503.07365},
  year={2025}
}

@article{wang2026vl,
  title={Vl-rethinker: Incentivizing self-reflection of vision-language models with reinforcement learning},
  author={Wang, Haozhe and Qu, Chao and Huang, Zuming and Chu, Wei and Lin, Fangzhen and Chen, Wenhu},
  journal={Advances in Neural Information Processing Systems},
  volume={38},
  pages={30865--30891},
  year={2026}
}

@inproceedings{goyal2017making,
  title={Making the v in vqa matter: Elevating the role of image understanding in visual question answering},
  author={Goyal, Yash and Khot, Tejas and Summers-Stay, Douglas and Batra, Dhruv and Parikh, Devi},
  booktitle={Proceedings of the IEEE conference on computer vision and pattern recognition},
  pages={6904--6913},
  year={2017}
}

@inproceedings{agrawal2018don,
  title={Don't just assume; look and answer: Overcoming priors for visual question answering},
  author={Agrawal, Aishwarya and Batra, Dhruv and Parikh, Devi and Kembhavi, Aniruddha},
  booktitle={Proceedings of the IEEE conference on computer vision and pattern recognition},
  pages={4971--4980},
  year={2018}
}

@inproceedings{niu2021counterfactual,
  title={Counterfactual vqa: A cause-effect look at language bias},
  author={Niu, Yulei and Tang, Kaihua and Zhang, Hanwang and Lu, Zhiwu and Hua, Xian-Sheng and Wen, Ji-Rong},
  booktitle={Proceedings of the IEEE/CVF conference on computer vision and pattern recognition},
  pages={12700--12710},
  year={2021}
}

@article{xia2025visionary,
  title={Visionary-r1: Mitigating shortcuts in visual reasoning with reinforcement learning},
  author={Xia, Jiaer and Zang, Yuhang and Gao, Peng and Li, Sharon and Zhou, Kaiyang},
  journal={arXiv preprint arXiv:2505.14677},
  year={2025}
}

@article{xiao2025perception,
  title={Perception-R1: Advancing Multimodal Reasoning Capabilities of MLLMs via Visual Perception Reward},
  author={Xiao, Tong and Xu, Xin and Huang, Zhenya and Gao, Hongyu and Liu, Quan and Liu, Qi and Chen, Enhong},
  journal={arXiv preprint arXiv:2506.07218},
  year={2025}
}

@inproceedings{zhang2025r1vl,
  title     = {R1-VL: Learning to Reason with Multimodal Large Language Models
               via Step-wise Group Relative Policy Optimization},
  author    = {Zhang, Jingyi and Huang, Jiaxing and Yao, Huanjin and
               Liu, Shunyu and Zhang, Xikun and Lu, Shijian and Tao, Dacheng},
  booktitle = {Proceedings of the IEEE/CVF International Conference on
               Computer Vision},
  pages     = {1859--1869},
  year      = {2025}
}

@article{wang2025perception,
  title={Perception-aware policy optimization for multimodal reasoning},
  author={Wang, Zhenhailong and Guo, Xuehang and Stoica, Sofia and Xu, Haiyang and Wang, Hongru and Ha, Hyeonjeong and Chen, Xiusi and Chen, Yangyi and Yan, Ming and Huang, Fei and others},
  journal={arXiv preprint arXiv:2507.06448},
  year={2025}
}

@article{huang2025spotlight,
  title={Spotlight on token perception for multimodal reinforcement learning},
  author={Huang, Siyuan and Qu, Xiaoye and Li, Yafu and Luo, Yun and He, Zefeng and Liu, Daizong and Cheng, Yu},
  journal={arXiv preprint arXiv:2510.09285},
  year={2025}
}

@article{wang2026visually,
  title={Visually-Guided Policy Optimization for Multimodal Reasoning},
  author={Wang, Zengbin and Xiong, Feng and Lin, Liang and Hu, Xuecai and Wang, Yong and Wang, Yanlin and Zhang, Man and Chu, Xiangxiang},
  journal={arXiv preprint arXiv:2604.09349},
  year={2026}
}

@article{li2026prpo,
  title={PRPO: Perception-Reinforced Policy Optimization via Token-Level Dynamic Advantage Reshaping},
  author={Li, Qiming and Li, Tianlun and Cheng, Xiaolong and Li, Hangyu and Gong, Ruiyan and Niu, Kangning and Jiang, Kaitao and Xu, Mu},
  journal={arXiv preprint arXiv:2606.08708},
  year={2026}
}

@article{liu2026noisyrollout,
  title={Noisyrollout: Reinforcing visual reasoning with data augmentation},
  author={Liu, Xiangyan and Ni, Jinjie and Wu, Zijian and Du, Chao and Dou, Longxu and Wang, Haonan and Pang, Tianyu and Shieh, Michael},
  journal={Advances in Neural Information Processing Systems},
  volume={38},
  pages={2923--2957},
  year={2026}
}

@article{li2025vision,
  title={Vision matters: Simple visual perturbations can boost multimodal math reasoning},
  author={Li, Yuting and Wei, Lai and Zheng, Kaipeng and Huang, Jingyuan and Kong, Linghe and Sun, Lichao and Huang, Weiran},
  journal={arXiv e-prints},
  pages={arXiv--2506},
  year={2025}
}

@article{zhang2025cf,
  title={Cf-vlm: Counterfactual vision-language fine-tuning},
  author={Zhang, Jusheng and Cai, Kaitong and Fan, Yijia and Wang, Jian and Wang, Keze},
  journal={arXiv preprint arXiv:2506.17267},
  year={2025}
}

@inproceedings{lu2021inter,
  title={Inter-gps: Interpretable geometry problem solving with formal language and symbolic reasoning},
  author={Lu, Pan and Gong, Ran and Jiang, Shibiao and Qiu, Liang and Huang, Siyuan and Liang, Xiaodan and Zhu, Song-Chun},
  booktitle={Proceedings of the 59th Annual Meeting of the Association for Computational Linguistics and the 11th International Joint Conference on Natural Language Processing (Volume 1: Long Papers)},
  pages={6774--6786},
  year={2021}
}

@inproceedings{lu2024mathvista,
  title={Mathvista: Evaluating mathematical reasoning of foundation models in visual contexts},
  author={Lu, Pan and Bansal, Hritik and Xia, Tony and Liu, Jiacheng and Li, Chunyuan and Hajishirzi, Hannaneh and Cheng, Hao and Chang, Kai-Wei and Galley, Michel and Gao, Jianfeng},
  booktitle={International Conference on Learning Representations},
  volume={2024},
  pages={23439--23554},
  year={2024}
}

@inproceedings{qiao2025wemath,
  title={We-math: Does your large multimodal model achieve human-like mathematical reasoning?},
  author={Qiao, Runqi and Tan, Qiuna and Dong, Guanting and MinhuiWu, MinhuiWu and Sun, Chong and Song, Xiaoshuai and Wang, Jiapeng and Gongque, Zhuoma and Lei, Shanglin and Zhang, Yifan and others},
  booktitle={Proceedings of the 63rd Annual Meeting of the Association for Computational Linguistics (Volume 1: Long Papers)},
  pages={20023--20070},
  year={2025}
}

@inproceedings{zhang2024mathverse,
  title={Mathverse: Does your multi-modal llm truly see the diagrams in visual math problems?},
  author={Zhang, Renrui and Jiang, Dongzhi and Zhang, Yichi and Lin, Haokun and Guo, Ziyu and Qiu, Pengshuo and Zhou, Aojun and Lu, Pan and Chang, Kai-Wei and Qiao, Yu and others},
  booktitle={European Conference on Computer Vision},
  pages={169--186},
  year={2024},
  organization={Springer}
}

@article{xiao2024logicvista,
  title={Logicvista: Multimodal llm logical reasoning benchmark in visual contexts},
  author={Xiao, Yijia and Sun, Edward and Liu, Tianyu and Wang, Wei},
  journal={arXiv preprint arXiv:2407.04973},
  year={2024}
}

@inproceedings{li2023super,
  title={Super-clevr: A virtual benchmark to diagnose domain robustness in visual reasoning},
  author={Li, Zhuowan and Wang, Xingrui and Stengel-Eskin, Elias and Kortylewski, Adam and Ma, Wufei and Van Durme, Benjamin and Yuille, Alan L},
  booktitle={Proceedings of the IEEE/CVF conference on computer vision and pattern recognition},
  pages={14963--14973},
  year={2023}
}

@inproceedings{yue2025mmmu,
  title={Mmmu-pro: A more robust multi-discipline multimodal understanding benchmark},
  author={Yue, Xiang and Zheng, Tianyu and Ni, Yuansheng and Wang, Yubo and Zhang, Kai and Tong, Shengbang and Sun, Yuxuan and Yu, Botao and Zhang, Ge and Sun, Huan and others},
  booktitle={Proceedings of the 63rd Annual Meeting of the Association for Computational Linguistics (Volume 1: Long Papers)},
  pages={15134--15186},
  year={2025}
}

@misc{openai2024gpt4o,
  author       = {{OpenAI}},
  title        = {Hello {GPT-4o}},
  year         = {2024},
  month        = may,
  howpublished = {OpenAI product announcement},
  url          = {https://openai.com/index/hello-gpt-4o/}
}

@misc{openai2024gpt4omini,
  author       = {{OpenAI}},
  title        = {{GPT-4o} mini: Advancing Cost-Efficient Intelligence},
  year         = {2024},
  month        = jul,
  howpublished = {OpenAI product announcement},
  url          = {https://openai.com/index/gpt-4o-mini-advancing-cost-efficient-intelligence/}
}

@misc{google2024gemini20,
  author       = {{Google DeepMind}},
  title        = {Introducing {Gemini 2.0}: Our New {AI} Model for the Agentic Era},
  year         = {2024},
  month        = dec,
  howpublished = {Google DeepMind blog},
  url          = {https://blog.google/innovation-and-ai/models-and-research/google-deepmind/google-gemini-ai-update-december-2024/}
}

@article{bai2025qwen25vl,
  title   = {{Qwen2.5-VL} Technical Report},
  author  = {Shuai Bai and Keqin Chen and Xuejing Liu and Jialin Wang
             and Wenbin Ge and Sibo Song and Kai Dang and Peng Wang
             and Shijie Wang and Jun Tang and Humen Zhong and Yuanzhi Zhu
             and Mingkun Yang and Zhaohai Li and Jianqiang Wan
             and Pengfei Wang and Wei Ding and Zheren Fu and Yiheng Xu
             and Jiabo Ye and Xi Zhang and Tianbao Xie and Zesen Cheng
             and Hang Zhang and Zhibo Yang and Haiyang Xu and Junyang Lin},
  journal = {arXiv preprint arXiv:2502.13923},
  year    = {2025}
}

@article{chen2024expanding,
  title={Expanding performance boundaries of open-source multimodal models with model, data, and test-time scaling},
  author={Chen, Zhe and Wang, Weiyun and Cao, Yue and Liu, Yangzhou and Gao, Zhangwei and Cui, Erfei and Zhu, Jinguo and Ye, Shenglong and Tian, Hao and Liu, Zhaoyang and others},
  journal={arXiv preprint arXiv:2412.05271},
  year={2024}
}

@article{li2024llava,
  title={Llava-onevision: Easy visual task transfer},
  author={Li, Bo and Zhang, Yuanhan and Guo, Dong and Zhang, Renrui and Li, Feng and Zhang, Hao and Zhang, Kaichen and Zhang, Peiyuan and Li, Yanwei and Liu, Ziwei and others},
  journal={arXiv preprint arXiv:2408.03326},
  year={2024}
}

@article{an2025llava,
  title={Llava-onevision-1.5: Fully open framework for democratized multimodal training},
  author={An, Xiang and Xie, Yin and Yang, Kaicheng and Zhang, Wenkang and Zhao, Xiuwei and Cheng, Zheng and Wang, Yirui and Xu, Songcen and Chen, Changrui and Zhu, Didi and others},
  journal={arXiv preprint arXiv:2509.23661},
  year={2025}
}

@article{wang2025llava,
  title={Llava-critic-r1: Your critic model is secretly a strong policy model},
  author={Wang, Xiyao and Li, Chunyuan and Yang, Jianwei and Zhang, Kai and Liu, Bo and Xiong, Tianyi and Huang, Furong},
  journal={arXiv preprint arXiv:2509.00676},
  year={2025}
}

@inproceedings{yang2025r1,
  title={R1-onevision: Advancing generalized multimodal reasoning through cross-modal formalization},
  author={Yang, Yi and He, Xiaoxuan and Pan, Hongkun and Jiang, Xiyan and Deng, Yan and Yang, Xingtao and Lu, Haoyu and Yin, Dacheng and Rao, Fengyun and Zhu, Minfeng and others},
  booktitle={Proceedings of the IEEE/CVF International Conference on Computer Vision},
  pages={2376--2385},
  year={2025}
}

@misc{gui2025training,
  title={Training reasoning model with dynamic advantage estimation on reinforcement learning},
  author={Gui, Lujun and Ren, Qingnan},
  year={2025}
}

@article{chen2025sft,
  title={Sft or rl? an early investigation into training r1-like reasoning large vision-language models},
  author={Chen, Hardy and Tu, Haoqin and Wang, Fali and Liu, Hui and Tang, Xianfeng and Du, Xinya and Zhou, Yuyin and Xie, Cihang},
  journal={arXiv preprint arXiv:2504.11468},
  year={2025}
}

@article{gou2025reasoning,
  title={Reasoning-Aligned Perception Decoupling for Scalable Multi-modal Reasoning},
  author={Gou, Yunhao and Chen, Kai and Liu, Zhili and Hong, Lanqing and Jin, Xin and Li, Zhenguo and Kwok, James T and Zhang, Yu},
  journal={arXiv preprint arXiv:2506.04559},
  year={2025}
}

@article{wang2026sota,
  title={Sota with less: Mcts-guided sample selection for data-efficient visual reasoning self-improvement},
  author={Wang, Xiyao and Yang, Zhengyuan and Feng, Chao and Lu, Hongjin and Li, Linjie and Lin, Chung-Ching and Lin, Kevin and Huang, Furong and Wang, Lijuan},
  journal={Advances in Neural Information Processing Systems},
  volume={38},
  pages={118818--118850},
  year={2026}
}

@article{yu2026dapo,
  title={Dapo: An open-source llm reinforcement learning system at scale},
  author={Yu, Qiying and Zhang, Zheng and Zhu, Ruofei and Yuan, Yufeng and Zuo, Xiaochen and Yue, Yu and Dai, Weinan and Fan, Tiantian and Liu, Gaohong and Liu, Lingjun and others},
  journal={Advances in Neural Information Processing Systems},
  volume={38},
  pages={113222--113244},
  year={2026}
}

@article{gao2026thinking,
  title={Thinking with Deltas: Incentivizing Reinforcement Learning via Differential Visual Reasoning Policy},
  author={Gao, Shujian and Wang, Yuan and Yan, Jiangtao and Wu, Zuxuan and Jiang, Yu-Gang},
  journal={arXiv preprint arXiv:2601.06801},
  year={2026}
}
\bibliographystyle{iclr2025_conference}

\appendix
\etocdepthtag.toc{appendix}
\clearpage


\etocsettagdepth{mainbody}{none}          
\etocsettagdepth{appendix}{subsubsection} 

\etocsettocstyle
  {\section*{Appendix Contents}%
   \noindent\rule{\linewidth}{0.4pt}\par\smallskip}
  {\par\smallskip\noindent\rule{\linewidth}{0.4pt}}

\etocsetnexttocdepth{subsubsection}
\tableofcontents
\clearpage

\clearpage

\section{Implementation Details}
\label{sec:impl_details}

\subsection{Training Recipe}
\label{sec:appendix_training_recipe}

The main text focuses on the SIVA-RL mechanism; this appendix reports the executable training recipe. Unless otherwise stated, all runs use Qwen2.5-VL backbones, AdamW-bf16, gradient checkpointing, same-image PatchSwap intervention views, and the reward function's answer-accuracy component as $r_{\mathrm{ans}}$. The answer reward is a scalar returned by the evaluator, not a strictly binary label. Training uses a two-stage schedule. Stage 1 is a 200-step PatchSwap perception-separation warm-up that treats all constructed intervention pairs as sensitivity-style perturbations without outcome routing or invariance alignment. Stage 2 runs 100 additional steps with SIVA-RL's paired outcome valuation and dual-direction routing enabled.

\begin{table}[t]
\centering
\caption{Core training and rollout hyperparameters for SIVA-RL.}
\label{tab:appendix_key_hyperparams}
\footnotesize
\setlength{\tabcolsep}{7pt}
\renewcommand{\arraystretch}{1.08}
\begin{tabular}{p{0.32\textwidth} p{0.58\textwidth}}
\toprule
\textbf{Hyperparameter} & \textbf{Value} \\
\midrule
\multicolumn{2}{c}{\textbf{General Training}} \\
\midrule
Backbone models & Qwen2.5-VL-3B and Qwen2.5-VL-7B \\
Optimizer / learning rate / weight decay & AdamW-bf16 / $10^{-6}$ / $10^{-2}$ \\
Maximum prompt / response length & $4096$ / $2048$ \\
Gradient checkpointing & enabled \\
Reward signal & answer-accuracy scalar with format and parse validity checks \\
\midrule
\multicolumn{2}{c}{\textbf{Rollout and Evaluation}} \\
\midrule
Rollouts per prompt & $G=5$ for training rollouts \\
Rollout temperature / top-$p$ & $1.0$ / $0.99$ \\
Intervention valuation decoding & sampled audit decoding with $M=4$ rollouts per view; temperature $0.7$, top-$p$ $0.95$ \\
Validation decoding multiplicity & GRPO: avg@8; DAPO: avg@8 \\
\midrule
\multicolumn{2}{c}{\textbf{Two-Stage Schedule}} \\
\midrule
Math stage 1 & ViRL39K train / MMK12 val; $4$ GPUs; $200$ max steps; PatchSwap warm-up without SIVA routing \\
Math stage 2 & ViRL39K train / MMK12 val; $4$ GPUs; $100$ max steps; dual-direction SIVA alignment \\
Medical stage 1 &  VQA-RAD SLAKE PathVQA  train split/ validation split; $4$ GPUs; $200$ max steps; PatchSwap warm-up without SIVA routing \\
Medical stage 2 & VQA-RAD SLAKE PathVQA  train split / validation split; $4$ GPUs; $100$ max steps; dual-direction SIVA alignment \\
\midrule
\multicolumn{2}{c}{\textbf{Batching}} \\
\midrule
Math global / rollout / validation batch & $128$ / $256$ in stage 1 and $512$ in stage 2 / $512$ \\
Medical global / rollout / validation batch & $64$ / $384$ / $256$ \\
Micro batch & $4$ for update and $8$ for experience computation \\
\midrule
\multicolumn{2}{c}{\textbf{Backbone-Specific RL Configuration}} \\
\midrule
GRPO-backbone & advantage estimator \texttt{grpo}; reference-KL branch enabled; online filtering off; default actor clipping \\
DAPO-backbone & advantage estimator \texttt{dapo}; reference-KL branch disabled; online filtering on with accuracy filter $[0.01,0.99]$; clip low $0.2$, clip high $0.28$ \\
KL penalty type / coefficient & low-variance KL / $10^{-2}$ \\
\bottomrule
\end{tabular}
\end{table}

\begin{table}[t]
\centering
\caption{PatchSwap and outcome-routing hyperparameters for SIVA-RL.}
\label{tab:appendix_siva_hyperparams}
\footnotesize
\setlength{\tabcolsep}{7pt}
\renewcommand{\arraystretch}{1.08}
\begin{tabular}{p{0.32\textwidth} p{0.58\textwidth}}
\toprule
\textbf{Hyperparameter} & \textbf{Value} \\
\midrule
\multicolumn{2}{c}{\textbf{PatchSwap Intervention Views}} \\
\midrule
Visual cell size & $28$ pixels, corresponding to a $14$-pixel vision patch with spatial merge size $2$ \\
Patch source & same-image non-candidate cells satisfying $d_{\min}$; if this source set is empty, or no source satisfies the distance constraint, we fall back to same-image cells excluding the target cell \\
Math candidate-cell ratio / replacement probability & $0.6$ / $0.75$ \\
Medical candidate-cell ratio / replacement probability & $0.4$ / $0.75$ \\
\midrule
\multicolumn{2}{c}{\textbf{Outcome Valuation and Alignment}} \\
\midrule
Clean correctness threshold $\tau_c$ & $0.75$  \\
Sensitivity reward-drop interval & $[\delta_{\mathrm{low}}^{\mathrm{sen}},\delta_{\mathrm{high}}^{\mathrm{sen}}]=[0.05,0.5]$ \\
Invariance reward-drop interval & $[\delta_{\mathrm{low}}^{\mathrm{inv}},\delta_{\mathrm{high}}^{\mathrm{inv}}]=[0.0,0.05]$ \\
Sensitivity coefficient / margin & $\lambda_{\mathrm{sen}}=0.005$, $m_{\mathrm{sen}}=0.05$ \\
Invariance coefficient / margin & $\lambda_{\mathrm{inv}}=0.001$, $m_{\mathrm{inv}}=0.01$ \\
\bottomrule
\end{tabular}
\end{table}

\subsection{Algorithm Pseudocode of SIVA-RL}
\label{sec:appendix_algorithm}

Algorithm~\ref{alg:siva_algorithm} summarizes one SIVA-RL training step. SIVA-RL differs from a standard grouped-rollout RL update in one way: it separates the clean rollout used for policy optimization from the paired audit used for intervention outcome valuation. The audit averages answer rewards over $M$ sampled rollouts and determines soft routing weights. The auxiliary loss then re-scores the already sampled clean responses under the clean and PatchSwap visual conditions.

\begin{algorithm}[t!]
\caption{SIVA-RL training step}
\label{alg:siva_algorithm}
\footnotesize
\begin{algorithmic}[1]
\Require Current policy $\pi_\theta$, old policy $\pi_{\theta_{\mathrm{old}}}$, audit policy $\pi_{\bar{\theta}}$, batch $\mathcal{B}=\{(q_i,x_i,a_i^\star)\}_{i=1}^{B}$
\Require Group size $G$, audit sample count $M$, PatchSwap parameters $(\eta,\rho,d_{\min})$, clean threshold $\tau_c$, reward-drop intervals, margins $(m_{\mathrm{sen}},m_{\mathrm{inv}})$, coefficients $(\lambda_{\mathrm{sen}},\lambda_{\mathrm{inv}})$
\Statex \textbf{Phase 1: Clean grouped rollout and base RL statistics}
\For{each prompt $(q_i,x_i)$ in $\mathcal{B}$}
    \State Sample clean responses $y_{i,g}\sim\pi_{\theta_{\mathrm{old}}}(\cdot\mid q_i,x_i)$ for $g=1,\ldots,G$
    \State Compute outcome rewards $R_{i,g}$ and grouped response advantages $A_{i,g}$
\EndFor
\Statex \textbf{Phase 2: Localized intervention construction}
\For{each image $x_i$}
    \State Sample candidate cells $C_i$ and effective targets $T_i$ on the visual-cell grid
    \State For each target cell, first sample a same-image non-candidate source cell with distance at least $d_{\min}$; if no valid source exists, fall back to all same-image cells except the target
    \State Construct PatchSwap intervention view $\tilde{x}_i$
\EndFor
\Statex \textbf{Phase 3: Outcome-conditioned valuation}
\For{each pair $(q_i,x_i,\tilde{x}_i)$}
    \State Set $x_i^{\mathrm{cl}}=x_i$ and $x_i^{\mathrm{ci}}=\tilde{x}_i$
    \For{$v\in\{\mathrm{cl},\mathrm{ci}\}$}
        \State Sample $\hat{y}_{i,m}^{v}\sim\pi_{\bar{\theta}}(\cdot\mid q_i,x_i^v)$ for $m=1,\ldots,M$ without gradient updates
        \State Compute $R_i^v=\frac{1}{M}\sum_{m=1}^{M}r_{\mathrm{ans}}(\hat{y}_{i,m}^{v},a_i^\star)$
        \State Compute parse-validity indicator $\Phi_i^{v}$
    \EndFor
    \State Compute validity gate $b_i=\mathbb{I}[R_i^{\mathrm{cl}}\ge\tau_c]\Phi_i^{\mathrm{cl}}\Phi_i^{\mathrm{ci}}$ and drop $\Delta_i=R_i^{\mathrm{cl}}-R_i^{\mathrm{ci}}$
    \State Convert $\Delta_i$ into soft routing weights $w_i^{\mathrm{sen}}$ and $w_i^{\mathrm{inv}}$
\EndFor
\Statex \textbf{Phase 4: Intervention-view re-scoring and alignment}
\For{each sampled response $y_{i,g}$}
    \State Re-score the same token sequence under $x_i^{\mathrm{cl}}$ and $x_i^{\mathrm{ci}}$
    \State Compute sampled divergence surrogate $D_{i,g}$
    \State Accumulate $w_i^{\mathrm{sen}}[m_{\mathrm{sen}}-D_{i,g}]_+$ and $w_i^{\mathrm{inv}}[D_{i,g}-m_{\mathrm{inv}}]_+$
\EndFor
\Statex \textbf{Phase 5: Policy update}
\State Update $\theta$ using $\mathcal{L}_{\mathrm{base}}+\lambda_{\mathrm{sen}}\mathcal{L}_{\mathrm{sen}}+\lambda_{\mathrm{inv}}\mathcal{L}_{\mathrm{inv}}$ and enabled backbone regularizers
\end{algorithmic}
\end{algorithm}

\subsection{Backbone Instantiations and Final Objectives}
\label{sec:appendix_backbone_objectives}

Let $\rho_{i,g,t}(\theta)$ denote the clean-rollout importance ratio defined in the main text, and let
\begin{equation}
\bar{\rho}_{i,g,t}(\theta)
=
\exp\!\Bigl(
\mathrm{clip}\bigl(
\log \rho_{i,g,t}(\theta),
\log(1-\epsilon_{\mathrm{low}}),
\log(1+\epsilon_{\mathrm{high}})
\bigr)
\Bigr)
\end{equation}
be the clipped ratio used by the shared actor-loss routine in our code. With response-level advantage $A_{i,g}$ broadcast to all tokens of response $y_{i,g}$, the per-token clipped actor loss can be written as
\begin{equation}
\ell^{\mathrm{PG}}_{i,g,t}
=
\begin{cases}
\max\!\left(
-A_{i,g}\rho_{i,g,t},
-A_{i,g}\bar{\rho}_{i,g,t}
\right), & A_{i,g}\ge 0,\\[1mm]
\min\!\left(
\max\!\left(
-A_{i,g}\rho_{i,g,t},
-A_{i,g}\bar{\rho}_{i,g,t}
\right),
-A_{i,g}\epsilon_{\mathrm{dual}}
\right), & A_{i,g}<0.
\end{cases}
\end{equation}
The clean-image policy-gradient term is then
\begin{equation}
\mathcal{L}_{\mathrm{PG}}(\theta)
=
\mathrm{Avg}_{i,g,t}\!\left[\ell^{\mathrm{PG}}_{i,g,t}\right].
\end{equation}
Here $\epsilon_{\mathrm{low}}$, $\epsilon_{\mathrm{high}}$, and $\epsilon_{\mathrm{dual}}$ denote the clipping hyperparameters used by the actor-loss routine.

In our implementation, the GRPO-backbone and DAPO-backbone share this same clipped actor-loss routine and differ primarily in the advantage estimator and associated training configuration. We therefore instantiate
\begin{equation}
\mathcal{L}^{\mathrm{G}}_{\mathrm{base}}(\theta)
=
\mathcal{L}_{\mathrm{PG}}(\theta;A^{\mathrm{grpo}})
+ \beta^{\mathrm{G}}_{\mathrm{ref}}\mathcal{L}_{\mathrm{KL\mbox{-}ref}}(\theta)
+ \beta^{\mathrm{G}}_{\mathrm{prcp}}\mathcal{L}_{\mathrm{prcp}}(\theta)
+ \beta^{\mathrm{G}}_{\mathrm{ori}}\mathcal{L}^{\mathrm{ori}}_{\mathrm{ent}}(\theta)
+ \beta^{\mathrm{G}}_{\mathrm{aug}}\mathcal{L}^{\mathrm{aug}}_{\mathrm{ent}}(\theta),
\end{equation}
and
\begin{equation}
\mathcal{L}^{\mathrm{D}}_{\mathrm{base}}(\theta)
=
\mathcal{L}_{\mathrm{PG}}(\theta;A^{\mathrm{dapo}})
+ \beta^{\mathrm{D}}_{\mathrm{ref}}\mathcal{L}_{\mathrm{KL\mbox{-}ref}}(\theta)
+ \beta^{\mathrm{D}}_{\mathrm{prcp}}\mathcal{L}_{\mathrm{prcp}}(\theta)
+ \beta^{\mathrm{D}}_{\mathrm{ori}}\mathcal{L}^{\mathrm{ori}}_{\mathrm{ent}}(\theta)
+ \beta^{\mathrm{D}}_{\mathrm{aug}}\mathcal{L}^{\mathrm{aug}}_{\mathrm{ent}}(\theta),
\end{equation}
where $\mathcal{L}_{\mathrm{prcp}}$ denotes an inherited perception-separation branch computed on the intervention-view condition, and disabled terms are omitted by setting their coefficients to zero according to the configuration. Stage-1 warm-up runs set \texttt{use\_vac\_papo=false}, so the full SIVA valuation loss and outcome-dependent routing stay inactive. The warm-up instead uses PatchSwap intervention views as sensitivity-style perturbations through the inherited perception-separation branch.

With SIVA enabled in stage 2, the final objectives become
\begin{equation}
\mathcal{L}^{\mathrm{G}}_{\mathrm{SIVA}}(\theta)
=
\mathcal{L}^{\mathrm{G}}_{\mathrm{base}}(\theta)
+ \lambda_{\mathrm{sen}}\mathcal{L}_{\mathrm{sen}}(\theta)
+ \lambda_{\mathrm{inv}}\mathcal{L}_{\mathrm{inv}}(\theta),
\end{equation}
and
\begin{equation}
\mathcal{L}^{\mathrm{D}}_{\mathrm{SIVA}}(\theta)
=
\mathcal{L}^{\mathrm{D}}_{\mathrm{base}}(\theta)
+ \lambda_{\mathrm{sen}}\mathcal{L}_{\mathrm{sen}}(\theta)
+ \lambda_{\mathrm{inv}}\mathcal{L}_{\mathrm{inv}}(\theta).
\end{equation}
The SIVA-specific alignment term is therefore shared across the two backbone families. Both reuse the same clean rollout responses, both compute the same outcome-conditioned sensitivity and low-drop consistency losses, and both differ only through the backbone configuration summarized in Tables~\ref{tab:appendix_key_hyperparams} and~\ref{tab:appendix_siva_hyperparams}.
Throughout the appendix, we keep the same notation as in the main text: $\lambda_{\mathrm{sen}}$ and $\lambda_{\mathrm{inv}}$ denote the sensitivity and invariance coefficients, $m_{\mathrm{sen}}$ and $m_{\mathrm{inv}}$ denote the corresponding margins, and $\mathcal{L}_{\mathrm{prcp}}$ refers only to the inherited PAPO perception branch rather than to a SIVA-specific term.

\clearpage
\section{Full Hyperparameter Sensitivity Results}
\label{sec:appendix_hparam}

\subsection{Sweep Setup and Full Results}
\label{sec:appendix_hparam_setup}

Table~\ref{tab:appendix_hparam_full} reports the complete per-benchmark sweeps
over the candidate-cell ratio $\eta$ and the minimum source distance
$d_{\min}$, the two knobs of the localized PatchSwap construction. We read
these sweeps as a test of \emph{where} and \emph{why} the construction
matters, not as a search for the best cell in the table.

All sweeps use the GRPO-3B backbone and the two-stage SIVA-RL schedule of
Appendix~\ref{sec:impl_details}. We isolate each construction hyperparameter
with a one-factor-at-a-time protocol. When we vary the candidate-cell ratio
$\eta$, the minimum source distance is fixed at its default $d_{\min}=0.25$.
When we vary $d_{\min}$, the candidate-cell ratio is fixed at $\eta=0.6$. All
remaining settings match the main configuration: the replacement probability
$\rho$, the grid resolution, the routing thresholds, the margins, and the loss
coefficients are held at the values in
Tables~\ref{tab:appendix_key_hyperparams}
and~\ref{tab:appendix_siva_hyperparams}. Each cell of
Table~\ref{tab:appendix_hparam_full} is therefore a full training run scored
under the same Avg@8 protocol as the main results. The numbers are directly
comparable to the GRPO-3B baseline of $47.92$.

We sweep $\eta$ and $d_{\min}$ because they control the two independent degrees
of freedom of a localized PatchSwap view: how much of the image is eligible for
replacement, and how far the replacement content is drawn from. The
candidate-cell ratio $\eta$ sets the spatial extent of the intervention. A
small $\eta$ perturbs a few cells, while a large $\eta$ exposes most of the
grid to replacement. The minimum source distance $d_{\min}$ sets the semantic
distance of the pasted content. A small $d_{\min}$ keeps the swap local and
visually plausible, while a large $d_{\min}$ imports distant, unrelated
structure. The replacement probability $\rho$ and the grid resolution interact
with both knobs, but they change the strength of the intervention rather than
its spatial or semantic character. We therefore hold $\rho$ and the grid fixed
and report their setting only through the main configuration.

\begin{table}[!htbp]
\centering
\caption{\textbf{Full per-benchmark hyperparameter ablations on Qwen2.5-VL-3B.} We vary one hyperparameter at a time while keeping the others fixed. All results are Avg@8 accuracies (\%).}
\vspace{1em}
\label{tab:appendix_hparam_full}
\fontsize{5.8pt}{6.4pt}\selectfont
\setlength{\tabcolsep}{2.5pt}
\renewcommand{\arraystretch}{1.02}
\resizebox{0.85\textwidth}{!}{%
\begin{tabular}{@{}lccccc@{\hspace{8pt}}ccccc@{}}
\toprule
\multirow{2}{*}{\textbf{Benchmark}}
& \multicolumn{5}{c}{\textbf{Candidate-cell-ratio sweep} $\boldsymbol{\eta}$}
& \multicolumn{5}{c}{\textbf{Minimum source-distance sweep} $\boldsymbol{d_{\min}}$} \\
\cmidrule(lr){2-6}\cmidrule(lr){7-11}
& \textbf{0.2} & \textbf{0.4} & \textbf{0.6} & \textbf{0.8} & \textbf{1.0}
& \textbf{0.00} & \textbf{0.25} & \textbf{0.35} & \textbf{0.75} & \textbf{0.90} \\
\midrule
\rowcolor{genblue}\multicolumn{11}{@{}l}{\textbf{General Reasoning}} \\
Geo3k & 32.57 & 33.13 & \textbf{34.26} & 32.30 & 33.69 & 33.96 & \textbf{34.26} & 34.01 & 32.82 & 33.88 \\
MathVista & 62.90 & 62.42 & 62.12 & \textbf{63.52} & 61.29 & 62.64 & 62.12 & \textbf{63.05} & 61.12 & 62.71 \\
We-Math & \textbf{63.00} & 62.28 & 61.92 & 60.94 & 62.69 & 61.28 & 61.92 & \textbf{62.05} & 60.83 & 61.82 \\
MMK12 & 60.56 & 59.77 & \textbf{64.63} & 60.67 & 59.97 & 60.38 & \textbf{64.63} & 61.52 & 60.52 & 59.55 \\
MathVerse & 58.47 & 58.90 & 59.66 & 57.63 & \textbf{60.09} & 59.00 & 59.66 & 59.68 & \textbf{59.74} & 59.58 \\
\rowcolor{avggray}\textbf{General AVG} & 55.50 & 55.30 & \textbf{56.52} & 55.01 & 55.55 & 55.45 & \textbf{56.52} & 56.06 & 55.01 & 55.51 \\
\addlinespace[1.5pt]
\rowcolor{visorange}\multicolumn{11}{@{}l}{\textbf{Vision-Dependent Reasoning}} \\
LogicVista & 41.08 & \textbf{42.67} & 42.42 & 40.80 & 41.58 & 42.20 & \textbf{42.42} & 41.95 & 42.03 & 40.97 \\
Counting & 64.25 & 67.12 & \textbf{75.19} & 68.19 & 75.12 & \textbf{75.38} & 75.19 & 67.69 & 71.12 & 68.94 \\
MMMU-Pro & 27.75 & 28.27 & \textbf{29.12} & 28.59 & 28.06 & 28.30 & \textbf{29.12} & 28.27 & 27.64 & 28.25 \\
MathVerse$_V$ & 55.44 & 55.86 & 55.68 & 54.66 & \textbf{56.61} & 55.66 & 55.68 & \textbf{57.10} & 56.16 & 56.14 \\
\rowcolor{avggray}\textbf{Vision AVG} & 47.13 & 48.48 & \textbf{50.60} & 48.06 & 50.34 & 50.38 & \textbf{50.60} & 48.75 & 49.24 & 48.57 \\
\midrule
\rowcolor{overallpurple}\textbf{Overall AVG} & 51.78 & 52.27 & \textbf{53.89} & 51.92 & 53.23 & 53.20 & \textbf{53.89} & 52.81 & 52.44 & 52.43 \\
\bottomrule
\end{tabular}}
\end{table}

Read column by column, Table~\ref{tab:appendix_hparam_full} shows that no
single benchmark drives the aggregate trends. The overall average stays within
a $2.1$-point band across the entire $\eta$ sweep and within a $1.5$-point band
across the $d_{\min}$ sweep. The per-benchmark optima are also dispersed: the
best $\eta$ for Geo3k, MathVista, and MathVerse differ from one another, and
the same holds for $d_{\min}$. The aggregate optimum at $\eta=0.6$,
$d_{\min}=0.25$ is therefore a robust compromise across benchmarks, not a
setting tuned to any one of them.

\subsection{Findings}
\label{sec:appendix_hparam_findings}

Three trends stand out.

\subsubsection{Both knobs have an interior optimum}
Neither hyperparameter improves monotonically. The candidate-cell ratio
$\eta$ peaks at the default $\eta=0.6$ ($53.89$ overall), drops to $51.92$ at
$\eta=0.8$, and then partially recovers to $53.23$ at $\eta=1.0$. The minimum
source distance $d_{\min}$ shows a cleaner interior peak: accuracy is highest
at $d_{\min}=0.25$ ($53.89$) and falls steadily as sources are forced farther
away ($52.81$, $52.44$, and $52.43$ at $0.35$, $0.75$, and $0.90$). Both
profiles follow from the construction rationale in
Section~\ref{sec:intervention_construction}: a useful intervention must be
hard enough to remove local visual evidence yet mild enough to keep the image
in-distribution. A too-small $d_{\min}$ copies a near-duplicate neighbor and
barely perturbs the evidence. A too-large $d_{\min}$ pastes an unrelated
distant patch and drifts toward arbitrary corruption. The intermediate setting
balances these two failure modes. This balanced regime is exactly what SIVA-RL
is designed to construct.

\subsubsection{Construction matters most where vision matters most}
Sensitivity to both knobs concentrates in the vision-dependent group. Across
the $\eta$ sweep, the vision-dependent average moves by $3.5$ points ($47.13$
to $50.60$), whereas the general average moves by only $1.5$ points ($55.01$
to $56.52$). The counting subset is the extreme case and swings by roughly
$11$ points ($64.25$ to $75.19$) as $\eta$ varies. General reasoning can lean
on textual structure and language priors, so it stays largely insensitive to
how the visual intervention is built. Vision-dependent reasoning has no such
fallback, so the construction directly governs the available learning signal.
This contrast corroborates the central premise of the paper. The visual
intervention becomes consequential when the task genuinely depends on visual
evidence. The pattern mirrors our main results, where SIVA-RL gains are
largest on the vision-dependent subset.

\subsubsection{Gains come from routing, not a finely tuned operator}
Every configuration in Table~\ref{tab:appendix_hparam_full} stays well above
the GRPO-3B baseline of $47.92$. Even the weakest setting ($51.78$ at
$\eta=0.2$) improves the overall average by $3.9$ points. Two configurations
are especially telling. The $\eta=1.0$ stress test, which falls back to the
relaxed source rule of Appendix~\ref{sec:impl_details}, still reaches $53.23$.
Removing the distance constraint entirely ($d_{\min}=0.00$) still reaches
$53.20$, close to the tuned $53.89$. SIVA-RL therefore does not depend on a
precisely calibrated perturbation operator. The construction hyperparameters
mainly control the \emph{supply} of informative clean--intervention pairs. The
outcome-conditioned routing then decides \emph{how} each pair is used.

\subsection{Discussion and Practical Recommendations}
\label{sec:appendix_hparam_discussion}

For practitioners, the sweep yields a simple recipe. Start from $\eta=0.6$ and
$d_{\min}=0.25$, the aggregate optimum, and treat both as coarse knobs rather
than quantities that require fine search. The flat interior of both profiles
means that any $\eta\in[0.4,0.6]$ and any $d_{\min}\in[0.25,0.35]$ lands within
roughly one point of the best overall average. A run that must minimize tuning
cost can fix both knobs at their defaults and still capture most of the gain.
This tolerance matters in multimodal RL, where a single configuration is
expensive and a dense grid search over construction hyperparameters is rarely
affordable.

The sweep also clarifies what the construction hyperparameters do and do not
control. They set the supply and difficulty of clean--intervention pairs, but
they do not decide how those pairs are used. That decision belongs to the
outcome-conditioned router of Section~\ref{sec:intervention_routing}. The
router reads the observed reward drop of each pair and assigns it to the
sensitivity branch, the invariance branch, or neither. The router absorbs much
of the variation that a fixed operator would otherwise propagate into training.
This absorption is why even the unconstrained and stress-test settings stay
close to the tuned optimum. Construction quality raises the ceiling of the
available signal. Routing determines how much of that ceiling the policy
reaches.

\clearpage
\section{Training-Dynamics Diagnostics}
\label{sec:appendix_training_dynamics}

We provide additional training-dynamics diagnostics for the math-domain runs. These plots are not used for model selection. They make the two-stage behavior of SIVA-RL more transparent.

\begin{figure}[!htbp]
    \centering
    \includegraphics[width=0.92\textwidth]{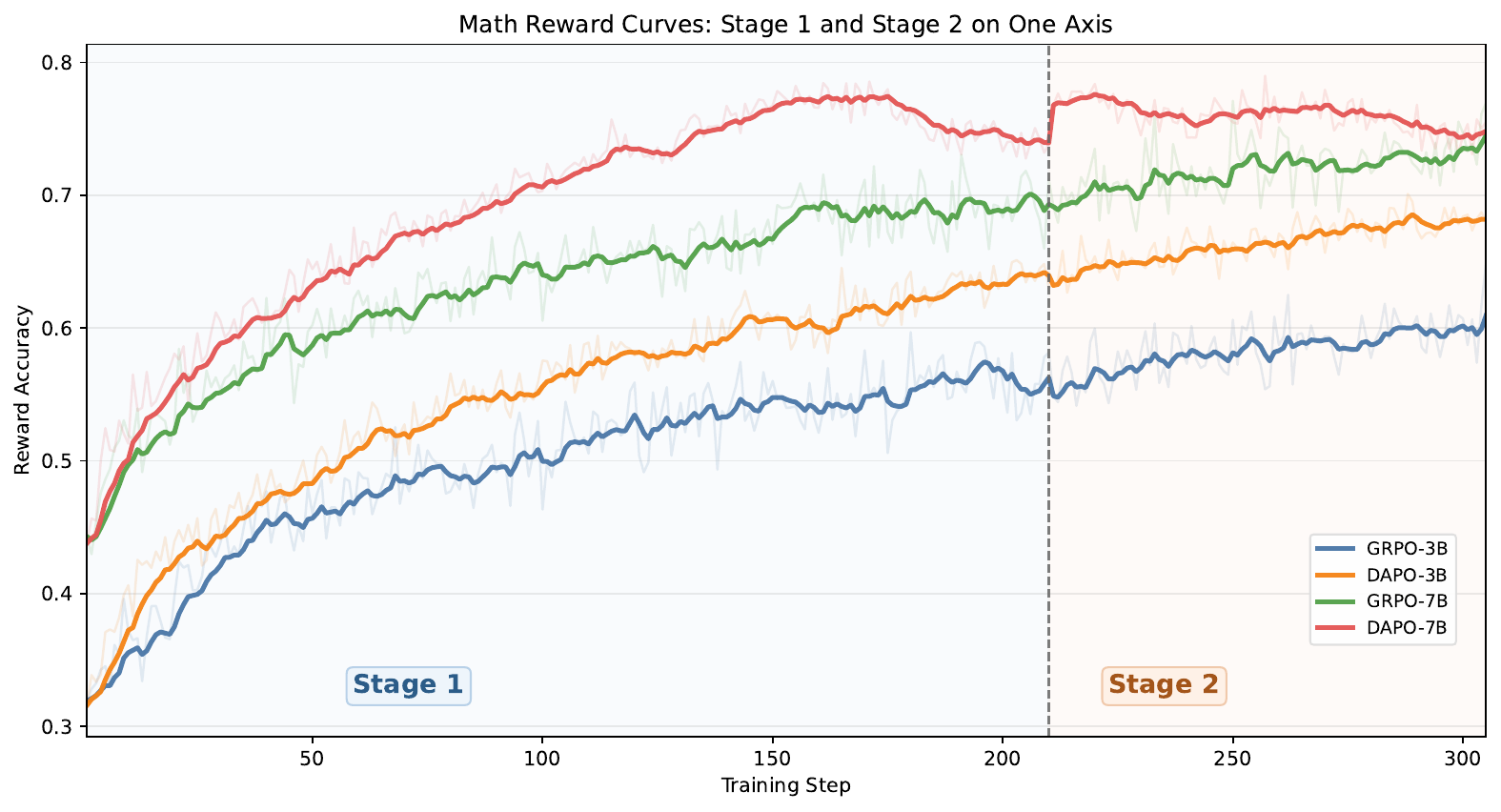}
    \caption{\textbf{Math-domain reward-accuracy dynamics across the two-stage schedule.} The dashed vertical line marks the transition from the 200-step PatchSwap/PAPO-style warm-up to the 100-step SIVA-RL stage, with curves shown for GRPO-3B, DAPO-3B, GRPO-7B, and DAPO-7B.}
    \label{fig:appendix_math_reward_overview}
\end{figure}

\subsection{Reward--Accuracy Dynamics}
\label{sec:appendix_reward_dynamics}

\noindent\textbf{The routing stage reignites a plateauing warm-up.}\quad
Figure~\ref{fig:appendix_math_reward_overview} preserves the expected scale and
backbone ordering throughout training: DAPO-7B leads, followed by GRPO-7B,
DAPO-3B, and GRPO-3B. During the 200-step warm-up, reward accuracy rises
quickly and then flattens as the uniform PatchSwap perturbation saturates. At
the transition to Stage 2 (dashed line), every curve resumes a clear upward
slope, and the larger models show a visible step up at the switch. The
continued improvement during the SIVA-RL stage indicates that
outcome-conditioned routing supplies a learning signal that the uniform warm-up
has already exhausted. This dynamic matches the stepwise ablation in
Section~\ref{sec:alignment_ablation}: adding routing and invariance alignment
on top of the warm-up raises the overall average from $52.64$ to $53.89$.

The backbone ordering is stable from the first steps to the last. The two 7B
models start near $0.44$ reward accuracy, while the two 3B models start near
$0.32$. This $0.10$--$0.13$ scale gap persists through both stages and never
closes. Within each scale, the DAPO backbone tracks above its GRPO counterpart
for most of training, so the four curves stay ordered as DAPO-7B, GRPO-7B,
DAPO-3B, and GRPO-3B without crossing.

The warm-up does more than flatten for the strongest model. DAPO-7B peaks near
$0.77$ around step $165$ and then drifts down to roughly $0.74$ by the
Stage-1/Stage-2 boundary, a mild regression under prolonged uniform
perturbation. The switch to outcome-conditioned routing reverses this drift.
DAPO-7B and GRPO-7B both jump discretely at the dashed line, and GRPO-7B then
climbs from about $0.69$ to $0.74$ across Stage 2. The 3B models gain more
gradually, while DAPO-7B holds near its post-jump level. Routing therefore
helps most where the warm-up had stalled or regressed, not uniformly across
configurations.

\begin{figure}[!htbp]
    \centering
    \includegraphics[width=0.92\textwidth]{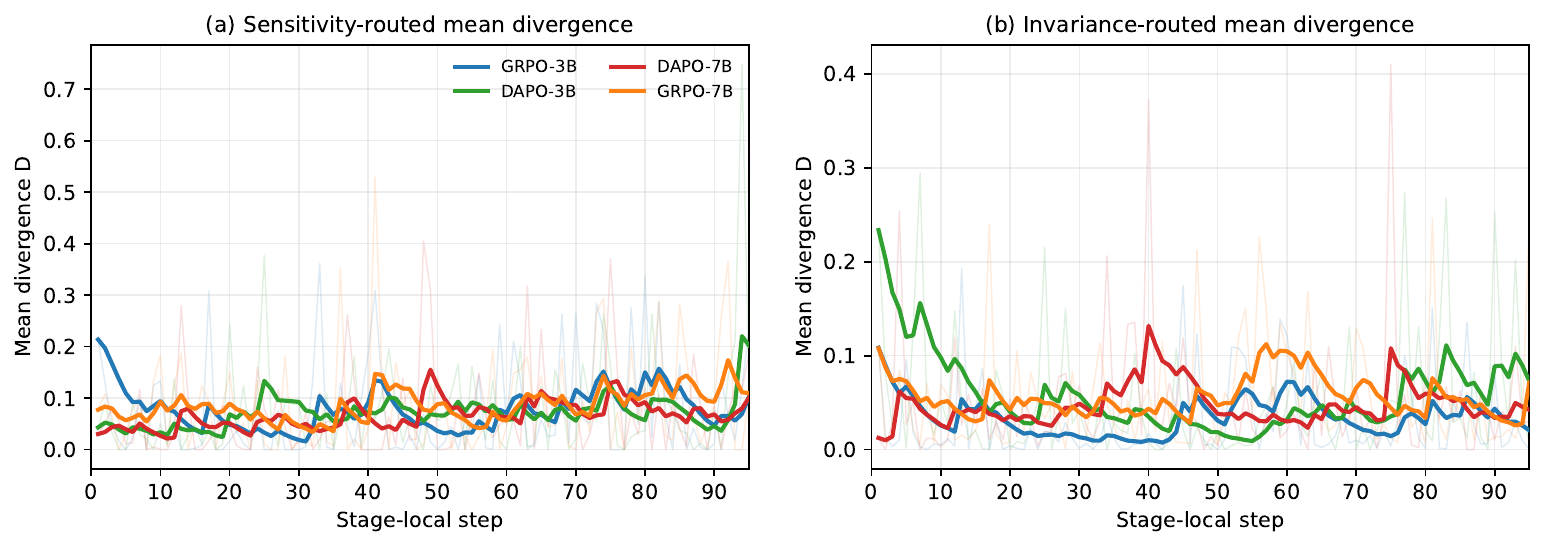}
    \caption{\textbf{intervention divergence dynamics during the math-domain SIVA-RL stage.} The two panels report mean divergence $D$ for sensitivity-routed and invariance-routed pairs, respectively. Each panel reports how the re-scored clean responses behave under PatchSwap interventions across backbone and model scales.}
    \label{fig:appendix_math_cf_divergence_trends}
\end{figure}

\subsection{Routing-Branch Separation}
\label{sec:appendix_routing_separation}

\noindent\textbf{Both routing branches stay active and separated.}\quad
Figure~\ref{fig:appendix_math_cf_divergence_trends} tracks the mean
clean-to-intervention divergence $D$ for the two routes during Stage 2. The
sensitivity branch holds $D$ at a positive, bounded level for all
configurations, consistent with the sensitivity margin $m_{\mathrm{sen}}$ that
pushes high-drop pairs apart. The invariance branch starts higher in the first
few steps and is then driven down toward the small invariance margin
$m_{\mathrm{inv}}$, where it stays for the rest of training. The two routes
never collapse into a single regime, and sensitivity divergence remains above
invariance divergence throughout. This separation is the training-time
signature of the dual-direction objective. The separation confirms the
mechanistic claim in Section~\ref{sec:mechanistic_analysis}: the same PatchSwap
operator keeps feeding both answer-changing and low-drop pairs rather than
degenerating into a uniform perturbation loss.

The two panels share a structure but sit at different scales; note the
different $y$-axis ranges. In the sensitivity panel, the smoothed curves stay
in a roughly $0.05$--$0.15$ band for all four configurations. They fluctuate
around the sensitivity margin rather than collapsing below it. In the
invariance panel, the transient at the start is the clearest signal. DAPO-3B
enters Stage 2 near $D=0.23$ and is pulled down within about fifteen steps to a
$0.03$--$0.05$ band, where the other configurations also settle. This early
descent is the invariance objective acting on low-drop pairs and tightening
clean-to-intervention consistency toward $m_{\mathrm{inv}}$.

After the transient, the two routes stay separated by a clear gap. Sensitivity
divergence holds around $0.05$--$0.15$ while invariance divergence stays around
$0.02$--$0.06$, so the sensitivity band sits roughly two to three times higher
throughout Stage 2. Late in training the sensitivity curves drift upward, with
GRPO-7B and DAPO-3B reaching about $0.2$ near the final steps. This upward
drift shows that PatchSwap keeps producing strongly answer-changing pairs
rather than running out of them. The faint unsmoothed traces spike as high as
$0.4$--$0.5$ in both panels. These bursts are individual high-divergence
batches that the soft routing weights absorb without destabilizing the smoothed
trend.

\clearpage
\section{Additional Medical-Domain Evaluation}
\label{sec:appendix_medical}

\subsection{Setup and Results}
\label{sec:appendix_medical_setup}

These results provide an additional cross-domain evaluation of the full training recipe. The comparison is made against the pretrained base model rather than a matched medical RL baseline, so the gains should not be attributed solely to the SIVA-RL auxiliary objective. For Qwen2.5-VL-3B (base) and Qwen2.5-VL-7B (base), the VQA-RAD, SLAKE, and PathVQA entries are our own Avg@8 evaluations of the official instruct checkpoints.

\begin{table}[!htbp]
   \centering
    \caption{\textbf{Full medical multimodal QA comparison across RL backbones and Qwen2.5-VL model scales.} Results are reported on the medical validation suite with Avg@8 evaluation.}
    \vspace{1em}
   \label{tab:medical_full_appendix}
   \small
   
   \setlength{\tabcolsep}{5pt}
   \renewcommand{\arraystretch}{1.08}
   \resizebox{\textwidth}{!}{
   \begin{tabular}{lccccc}
   \toprule
   \multirow{2}{*}{\textbf{Method}} 
   & \multicolumn{3}{c}{\cellcolor{blue!10}\textbf{In-Domain Datasets}}
   & \multicolumn{1}{c}{\cellcolor{orange!12}\textbf{Out-of-Domain Datasets}}
   & \multicolumn{1}{c}{\cellcolor{gray!12}\textbf{Overall}} \\
   \cmidrule(lr){2-4} \cmidrule(lr){5-5} \cmidrule(lr){6-6}
   & \textbf{VQA-RAD} 
   & \textbf{SLAKE} 
   & \textbf{PathVQA} 
   & \textbf{MedXpertQA} 
   & \textbf{AVG} \\
   \midrule
   
   \multicolumn{6}{c}{\textbf{Proprietary Models}} \\
   \midrule
   Gemini-2.0-flash-lite & 59.4 & 73.1 & 64.9 & -- & -- \\
   GPT-4.1-Nano          & 61.8 & 73.1 & 70.6 & -- & -- \\
   GPT-4o                & 63.9 & 71.6 & 75.9 & -- & -- \\
   
   \midrule
   \multicolumn{6}{c}{\textbf{General-Purpose Multimodal VLMs}} \\
   \midrule
   Qwen-VL-Chat        & 47.0 & 56.0 & 55.1 & -- & -- \\
   Yi-VL-34B              & 53.0 & 58.9 & 47.3 & -- & -- \\
   LLaVA-v1.6-7B        & 52.6 & 57.9 & 47.9 & -- & -- \\
   LLaVA-v1.6-13B       & 55.8 & 58.9 & 51.9 & -- & -- \\
   LLaVA-v1.6-34B       & 58.6 & 67.3 & 59.1 & -- & -- \\
   LLaVA-v1.5-LLaMA3-8B 
                                           & 54.2 & 59.4 & 54.1 & -- & -- \\
   
   \midrule
   \multicolumn{6}{c}{\textbf{Medical Multimodal VLMs}} \\
   \midrule
   Med-Flamingo          & 45.4 & 43.5 & 54.7 & 22.1 & 41.4 \\
   RadFM                   & 50.6 & 34.6 & 38.7 & \underline{23.4} & 36.8 \\
   LLaVA-Med-7B            & 51.4 & 48.6 & 56.8 & 20.8 & 44.4 \\
   LLaVA\_Med-LLaMA3-8B
                                           & 60.2 & 61.2 & 54.5 & -- & -- \\
   PubMedVision-8B      & 63.8 & 74.5 & 59.9 & -- & -- \\
   HuatuoGPT-Vision-34B & 68.1 & 76.9 & 63.5 & 22.1 & 57.7 \\
   MedVLThinker-7B     & 63.7 & 67.8 & 65.2 & 20.9 & 54.4 \\
   CAPO-7B              & \textbf{78.5} & \underline{79.1} & 68.9 & -- & -- \\
   
   \midrule
   \multicolumn{6}{c}{\textbf{Medical Agentic Systems}} \\
   \midrule
   MedAgents             & 65.6 & 67.9 & 63.2 & -- & -- \\
   MDAgents               & 66.8 & 68.2 & 65.4 & -- & -- \\
   AFlow               & 67.3 & 68.9 & 66.4 & -- & -- \\
   MMedAgent-RL-7B        & \underline{71.5} & 76.2 & 72.3 & -- & -- \\

   \midrule
   \multicolumn{6}{c}{\textbf{Our Models}} \\
   \midrule
   Qwen2.5-VL-3B (base) & 54.6 & 57.9 & 49.2 & 20.7 & 45.6 \\
   \hspace{2em}DAPO + SIVA-RL & \gainds{62.7}{+8.1} & \gainds{68.1}{+10.2} & \gainds{66.5}{+17.3} & \gainds{23.0}{+2.3} & \gainds{55.1}{+9.5} \\
   \hspace{2em}GRPO + SIVA-RL & \gainds{63.0}{+8.4} & \gainds{68.8}{+10.9} & \gainds{65.2}{+16.0} & \gaindbs{24.5}{+3.8} & \gainds{55.4}{+9.8} \\
   Qwen2.5-VL-7B (base) & 62.1 & 62.6 & 54.9 & 20.1 & 49.9 \\
   \hspace{2em}GRPO + SIVA-RL & 67.1{\scriptsize\textcolor{green!60!black}{\,(+5.0)}} & \textbf{79.9}{\scriptsize\textcolor{green!60!black}{\,(+17.3)}} & \textbf{80.7}{\scriptsize\textcolor{green!60!black}{\,(+25.8)}} & 22.3{\scriptsize\textcolor{green!60!black}{\,(+2.2)}} & \textbf{62.5}{\scriptsize\textcolor{green!60!black}{\,(+12.6)}} \\
   \hspace{2em}DAPO + SIVA-RL & 67.4{\scriptsize\textcolor{green!60!black}{\,(+5.3)}} & 72.7{\scriptsize\textcolor{green!60!black}{\,(+10.1)}} & \underline{77.4}{\scriptsize\textcolor{green!60!black}{\,(+22.5)}} & 22.3{\scriptsize\textcolor{green!60!black}{\,(+2.2)}} & \underline{60.0}{\scriptsize\textcolor{green!60!black}{\,(+10.1)}} \\
   
   \bottomrule
   \end{tabular}
   }

   \label{tab:medical_full_appendix}
\end{table}

\subsection{Grounding Caveat}
\label{sec:appendix_medical_caveat}

Medical QA accuracy alone is not evidence of improved visual grounding.
Medical benchmarks often combine image findings with clinical text, pathology,
laboratory information, and diagnosis labels, so answer gains may reflect
improved use of non-visual cues. A stricter evidence-level evaluation would
separate visually observable findings from non-visual evidence and check
whether gains persist under text-only, blank-image, and leak-aware controls.
We therefore treat the medical results above as cross-domain performance
evidence, not a standalone proof of medical visual grounding.

\section{Qualitative Case Studies}
\label{sec:appendix_qualitative}
We present qualitative cases to clarify what the quantitative gains in the main paper mean at the sample level. The cases are organized around diagnostic roles rather than visual appeal. They fall into five categories:
\begin{enumerate}
    \item Visual recovery
    \item Misleading black-image correctness
    \item PatchSwap-only instability
    \item Local geometric stability
    \item Residual failure
\end{enumerate}

This structure matters because SIVA-RL does not assume that every visual intervention induces the same training signal. SIVA-RL instead interprets a perturbation through the paired outcome change: large drops provide sensitivity supervision, low-drop pairs provide invariance supervision, and persistent failures reveal limitations outside the current training signal.

On the diagnostic benchmark probes, random mask, PatchSwap, and black-image interventions all produce mixtures of outcome-drop and outcome-stable behavior. The perturbation operator alone therefore cannot determine whether a paired example should be separated or kept consistent. The examples below illustrate this heterogeneity and show one failure mode that SIVA-RL does not eliminate. They clarify the limitations of SIVA-RL and identify directions for future work.

\newcommand{\caseimage}[3]{%
  \begin{minipage}[t]{#1}\centering
    \includegraphics[width=\linewidth]{#2}\\[-1mm]{\scriptsize #3}%
  \end{minipage}}

\newcommand{\caseoutcomes}[6]{%
\begin{tabular}{lccc}
\toprule
 & Clean & Black & PatchSwap \\ \midrule
GRPO    & #1 & #2 & #3 \\
SIVA-RL & #4 & #5 & #6 \\
\bottomrule
\end{tabular}}

\tcbset{
  caseouter/.style={colback=gray!4,colframe=gray!40,breakable,
    width=\textwidth,fonttitle=\bfseries\normalsize,
    boxrule=0.8pt,top=4pt,bottom=4pt,left=4pt,right=4pt},
  casetask/.style={colback=blue!5,colframe=blue!30,breakable,
    boxrule=0.4pt,fonttitle=\bfseries\small,top=2pt,bottom=2pt},
  casebaseline/.style={colback=red!5,colframe=red!35,breakable,
    boxrule=0.4pt,fonttitle=\bfseries\small,top=2pt,bottom=2pt},
  casesiva/.style={colback=green!5,colframe=green!40,breakable,
    boxrule=0.4pt,fonttitle=\bfseries\small,top=2pt,bottom=2pt},
  caseinterp/.style={colback=gray!6,colframe=gray!35,breakable,
    boxrule=0.4pt,fonttitle=\bfseries\small,top=2pt,bottom=2pt},
}

\clearpage
\subsection{Visual Recovery}
\label{sec:case_visual_recovery}
\begin{tcolorbox}[caseouter,title={Case 1 — Visual Recovery}]
\footnotesize

\noindent
\begin{minipage}[t]{0.44\textwidth}
\centering
\caseimage{0.62\linewidth}{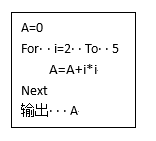}{clean image}
\end{minipage}\hfill
\begin{minipage}[t]{0.52\textwidth}
\begin{tcolorbox}[casetask,title=Task]
\textbf{Benchmark.} MMK12, data\_343. \quad \textbf{Gold.} D.\\
\textbf{Question.} After running the visual program, what is the output value?
\end{tcolorbox}
\end{minipage}

\medskip

\noindent
\begin{minipage}[t]{0.48\textwidth}
\begin{tcolorbox}[casebaseline,title=GRPO Baseline — wrong visual reading]
\textbf{Prediction.} B on the clean image.\\
\textbf{Saved response excerpt.} The model reads the update as ``$A=A+i^2\cdot i$'' and then accumulates an incorrect sequence.
\end{tcolorbox}
\end{minipage}\hfill
\begin{minipage}[t]{0.48\textwidth}
\begin{tcolorbox}[casesiva,title=SIVA-RL — recovers the rendered loop]
\textbf{Prediction.} D on the clean image.\\
\textbf{Saved response excerpt.} The model follows the loop from $i=2$ to $i=5$ with ``$A=A+i\cdot i$'', producing the correct option.
\end{tcolorbox}
\end{minipage}

\medskip

\begin{tcolorbox}[caseinterp,title=Interpretation]
\textbf{Outcome pattern.} \caseoutcomes{\textcolor{red}{B}}{\textcolor{red}{C}}{\textcolor{red}{A}}{\textcolor{green!45!black}{D}}{\textcolor{red}{C}}{\textcolor{green!45!black}{D}}\\[1mm]
This case illustrates visual recovery rather than memorized answer correction. The black-image probe remains incorrect for SIVA-RL, so the correct clean prediction cannot be attributed to a text-only prior. The PatchSwap result stays correct because the local replacement does not destroy the program structure needed to read the loop. In SIVA-RL terms, this behavior is exactly the goal: preserve the response under low-impact local perturbations while still relying on the visible program when the image is available.
\end{tcolorbox}

\end{tcolorbox}
\captionof{figure}{\textbf{Visual recovery case.} SIVA-RL corrects a baseline OCR/program-reading error by using the rendered loop, while the black-image response remains incorrect.}
\label{fig:case_program_ocr_recovery}

\clearpage 
\subsection{Misleading Black-Image Correctness}
\label{sec:case_black_spurious}
\begin{tcolorbox}[caseouter,title={Case 2 — Misleading Black-Image Correctness}]
\footnotesize

\noindent
\begin{minipage}[t]{0.57\textwidth}
\caseimage{0.48\linewidth}{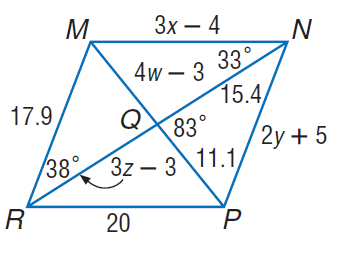}{clean image}\hfill
\caseimage{0.48\linewidth}{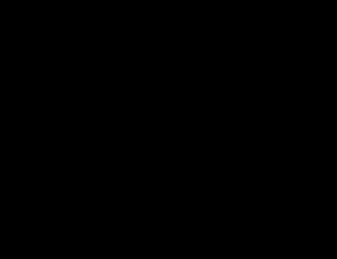}{black-image probe}
\end{minipage}\hfill
\begin{minipage}[t]{0.39\textwidth}
\begin{tcolorbox}[casetask,title=Task]
\textbf{Benchmark.} AI4Math\_MathVerse, AI4Math\_MathVerse\_1328. \quad \textbf{Gold.} D.\\
\textbf{Question.} Use parallelogram $MNPR$ to find $m\angle RMN$. Choices: A:33, B:38, C:71, D:109.
\end{tcolorbox}
\end{minipage}

\medskip

\noindent
\begin{minipage}[t]{0.48\textwidth}
\begin{tcolorbox}[casebaseline,title=GRPO Baseline — black-correct but clean-failed]
\textbf{Predictions.} Clean: no parse; black: D.\\
\textbf{Saved response excerpt.} On the clean diagram, it applies generic parallelogram facts but does not produce a valid final answer; on the black image, it guesses the correct option from text alone.
\end{tcolorbox}
\end{minipage}\hfill
\begin{minipage}[t]{0.48\textwidth}
\begin{tcolorbox}[casesiva,title=SIVA-RL — correct only with visible geometry]
\textbf{Predictions.} Clean: D; black: B.\\
\textbf{Saved response excerpt.} The clean response uses adjacent/supplementary angle constraints in the visible parallelogram and selects D.
\end{tcolorbox}
\end{minipage}

\medskip

\begin{tcolorbox}[caseinterp,title=Interpretation]
This case shows why black-image correctness is not reliable evidence of invariance. The baseline becomes correct only after the diagram is removed. This behavior is more consistent with answer-prior guessing than with robust visual reasoning. Treating such a pair as positive invariance would reward visually unsupported behavior. SIVA-RL avoids this failure mode by conditioning the auxiliary direction on paired outcomes and validity checks rather than on the intervention identity alone.
\end{tcolorbox}

\end{tcolorbox}
\captionof{figure}{\textbf{Misleading black-image correctness.} The baseline's correct black-image answer is not evidence of robust visual reasoning; it masks a clean-image failure.}
\label{fig:case_parallelogram_black_spurious}

\clearpage

\subsection{PatchSwap-Only Instability (Chart)}
\label{sec:case_patchswap_chart}
\begin{tcolorbox}[caseouter,title={Case 3 — PatchSwap Diagnostic (Chart)}]
\footnotesize

\noindent
\begin{minipage}[t]{0.57\textwidth}
\caseimage{0.48\linewidth}{}{clean image}\hfill
\caseimage{0.48\linewidth}{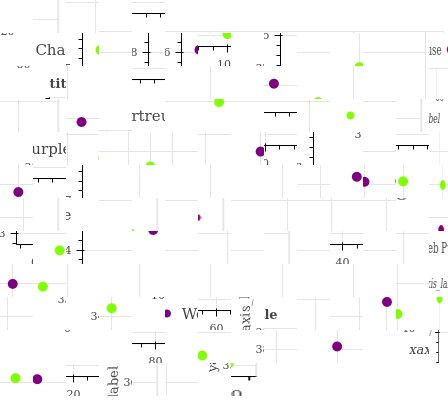}{PatchSwap intervention}
\end{minipage}\hfill
\begin{minipage}[t]{0.39\textwidth}
\begin{tcolorbox}[casetask,title=Task]
\textbf{Benchmark.} AI4Math\_MathVista, AI4Math\_MathVista\_324. \quad \textbf{Gold.} no.\\
\textbf{Question.} Does Web Purple have the maximum area under the curve?
\end{tcolorbox}
\end{minipage}

\medskip

\noindent
\begin{minipage}[t]{0.48\textwidth}
\begin{tcolorbox}[casebaseline,title=GRPO Baseline — PatchSwap-only failure]
\textbf{Predictions.} Clean: no; PatchSwap: yes.\\
\textbf{Saved response excerpt.} The clean response compares Chartreuse and Web Purple correctly, but the PatchSwap response concludes that Web Purple is higher.
\end{tcolorbox}
\end{minipage}\hfill
\begin{minipage}[t]{0.48\textwidth}
\begin{tcolorbox}[casesiva,title=SIVA-RL — stable under local replacement]
\textbf{Predictions.} Clean: no; PatchSwap: no.\\
\textbf{Saved response excerpt.} The response continues to compare the curve values and area under the localized intervention, keeping the correct answer.
\end{tcolorbox}
\end{minipage}

\medskip

\begin{tcolorbox}[caseinterp,title=Interpretation]
Both models are correct on the clean image and on the black probe. Only PatchSwap exposes the baseline instability. This result supports the need for distance-constrained local interventions rather than only full-image removal.\\[2pt]
The baseline appears reliable under clean and black-image predictions, but the localized intervention changes the answer. This case is exactly where SIVA-RL's PatchSwap construction is useful: it probes whether the model's decision is stable to irrelevant local changes while preserving much of the original visual context. The SIVA-RL response remains correct under PatchSwap. This stability suggests that the low-drop consistency signal can suppress brittle local visual associations without forcing invariance to destructive full-image removal.
\end{tcolorbox}

\end{tcolorbox}
\captionof{figure}{\textbf{PatchSwap diagnostic case.} A localized within-image replacement reveals a baseline stability gap that the black-image probe does not reveal.}
\label{fig:case_curve_area_patchswap_gap}

\clearpage

\subsection{Local Geometric Stability}
\label{sec:case_patchswap_geometry}
\begin{tcolorbox}[caseouter,title={Case 4 — PatchSwap Diagnostic (Geometry)}]
\footnotesize

\noindent
\begin{minipage}[t]{0.57\textwidth}
\caseimage{0.48\linewidth}{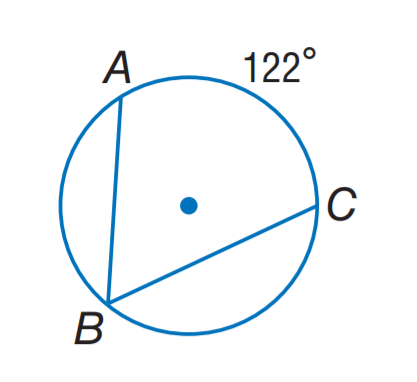}{clean image}\hfill
\caseimage{0.48\linewidth}{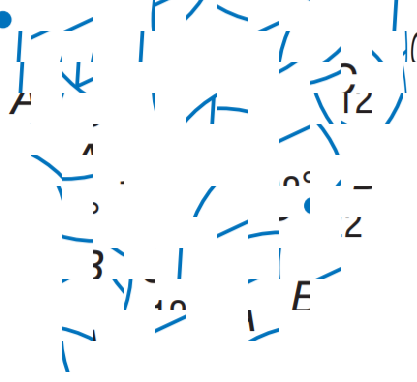}{PatchSwap intervention}
\end{minipage}\hfill
\begin{minipage}[t]{0.39\textwidth}
\begin{tcolorbox}[casetask,title=Task]
\textbf{Benchmark.} AI4Math\_MathVerse, AI4Math\_MathVerse\_1331. \quad \textbf{Gold.} C.\\
\textbf{Question.} Find $m\angle B$. Choices: A:29, B:58, C:61, D:122.
\end{tcolorbox}
\end{minipage}

\medskip

\noindent
\begin{minipage}[t]{0.48\textwidth}
\begin{tcolorbox}[casebaseline,title=GRPO Baseline — local geometry break]
\textbf{Predictions.} Clean: C; PatchSwap: B.\\
\textbf{Saved response excerpt.} The clean response uses the inscribed-angle relation, but the PatchSwap response treats the figure as a generic triangle and switches to B.
\end{tcolorbox}
\end{minipage}\hfill
\begin{minipage}[t]{0.48\textwidth}
\begin{tcolorbox}[casesiva,title=SIVA-RL — consistent final decision]
\textbf{Predictions.} Clean: C; PatchSwap: C.\\
\textbf{Saved response excerpt.} The model keeps the correct final option under the localized intervention instead of changing to the triangle-based answer.
\end{tcolorbox}
\end{minipage}

\medskip

\begin{tcolorbox}[caseinterp,title=Interpretation]
This additional geometry case mirrors the chart case in a symbolic diagram. Clean accuracy alone does not reveal whether the model has learned a stable geometric relation or merely followed a fragile visual cue. PatchSwap exposes the baseline's dependence on local diagram fragments: the answer changes from the correct arc/angle relation to a generic triangle answer. SIVA-RL keeps the decision fixed. This stability is the intended invariance behavior when the perturbation does not change the task-relevant relation.
\end{tcolorbox}

\end{tcolorbox}
\captionof{figure}{\textbf{Additional PatchSwap case on geometry.} The localized intervention uncovers a baseline error that is absent on the clean image.}
\label{fig:case_circle_arc_patchswap_gap}

\clearpage

\subsection{Residual Failure: Dense Multi-Instance Counting}
\label{sec:case_counting_failure}
\begin{tcolorbox}[caseouter,title={Case 5 — Residual Failure: Dense Multi-Instance Counting}]
\footnotesize

\noindent
\begin{minipage}[t]{0.64\textwidth}
\caseimage{0.32\linewidth}{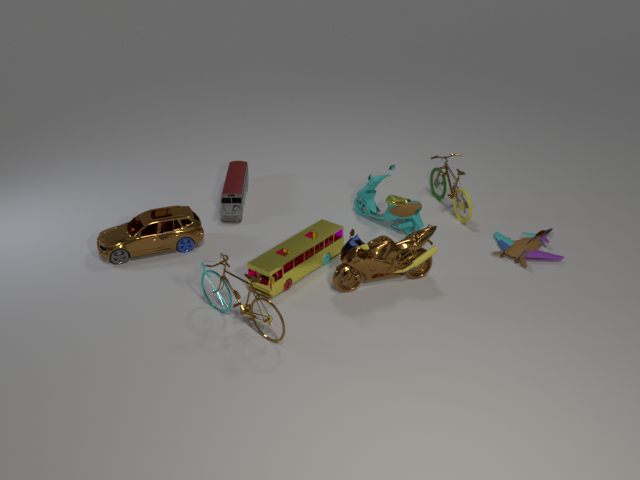}{clean image}\hfill
\caseimage{0.32\linewidth}{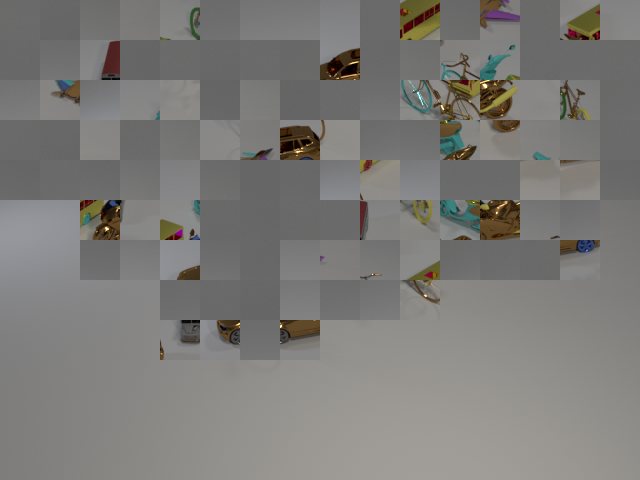}{PatchSwap intervention}\hfill
\caseimage{0.32\linewidth}{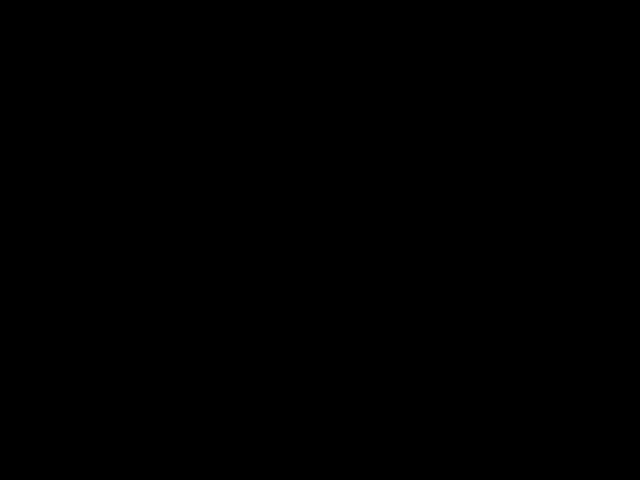}{black-image probe}
\end{minipage}\hfill
\begin{minipage}[t]{0.32\textwidth}
\begin{tcolorbox}[casetask,title=Task]
\textbf{Benchmark.} BUAADreamer\_clevr\_count, sample 62. \quad \textbf{Gold.} 8.\\
\textbf{Question.} How many different items are there in the image?
\end{tcolorbox}
\end{minipage}

\medskip

\noindent
\begin{minipage}[t]{0.48\textwidth}
\begin{tcolorbox}[casesiva,title=SIVA-RL Clean — near miss]
\textbf{Prediction.} 9 on the clean image.\\
\textbf{Saved response excerpt.} The model enumerates vehicle-like objects but double-counts or over-separates visually similar instances, producing one extra item.
\end{tcolorbox}
\end{minipage}\hfill
\begin{minipage}[t]{0.48\textwidth}
\begin{tcolorbox}[casebaseline,title=Perturbed Views — unstable counts]
\textbf{Predictions.} PatchSwap: 15; black: 0; random mask: 20.\\
\textbf{Saved response excerpt.} Under PatchSwap and random mask, the model falls back to a generic item-counting strategy and over-counts fragmented or repeated visual evidence.
\end{tcolorbox}
\end{minipage}

\begin{tcolorbox}[caseinterp,title=Interpretation]
This failure case marks a boundary of the current method. SIVA-RL improves whether the policy reacts to visual evidence, but it does not guarantee perfect object individuation in dense scenes with overlapping, visually similar instances. The clean prediction is already wrong, and all three perturbed views remain wrong, so the example provides little reliable positive supervision for either sensitivity or invariance. Such persistent-error cases motivate stronger object-level counting supervision or region-level credit assignment beyond the current answer-level paired-outcome signal.
\end{tcolorbox}

\end{tcolorbox}
\captionof{figure}{\textbf{Residual failure on dense multi-instance counting.} Even after SIVA-RL, the model can over-count similar objects on the clean image and become more unstable under local or global visual perturbations.}
\label{fig:case_counting_failure}

\end{document}